\documentclass{article}

\usepackage[dblblindworkshop, final]{neurips_2026}
\workshoptitle{AI for Drug Discovery: Bridging the Translation Gap}
\usepackage[utf8]{inputenc} % allow utf-8 input
\usepackage[T1]{fontenc}    % use 8-bit T1 fonts
\usepackage{hyperref}       % hyperlinks
\usepackage{url}            % simple URL typesetting
\usepackage{booktabs}       % professional-quality tables
\usepackage{amsfonts}       % blackboard math symbols
\usepackage{nicefrac}       % compact symbols for 1/2, etc.
\usepackage{microtype}      % microtypography
\usepackage{xcolor}         % colors
\usepackage{multirow}
\usepackage{graphicx}
\usepackage{tabularx}
\usepackage{seqsplit}
\usepackage{listings}
\usepackage{adjustbox}
\usepackage{makecell}
\usepackage{hyperref}

\hypersetup{hidelinks}

\title{Toward a Locally Deployable Agentic Co-Scientist: Small-Model Planning for Early-Stage Drug Discovery}

\author{
\makebox[0pt][c]{%
Tian Liang$^{1,2,4}$ \hspace{0.8em}
Jiayu Chang$^{5}$ \hspace{0.8em}
Alejandro F. Frangi$^{2,3,4,6,7}$ \hspace{0.8em}
Mobarak I. Hoque$^{3,4,6, *}$ \hspace{0.8em}
Richard A. Bryce$^{1, *}$%
}
\\[6pt]
\makebox[0pt][c]{%
$^{1}$ Division of Pharmacy and Optometry, University of Manchester, UK%
}
\\
\makebox[0pt][c]{%
$^{2}$ Department of Computer Science, University of Manchester, UK%
}
\\
\makebox[0pt][c]{%
$^{3}$ Division of Informatics, Imaging and Data Science, University of Manchester, UK%
}
\\
\makebox[0pt][c]{%
$^{4}$ Centre for Computational Imaging and Modelling in Medicine (CIMIM), University of Manchester, UK%
}
\\
\makebox[0pt][c]{%
$^{5}$ Department of Electrical Engineering, Stanford University, USA%
}
\\
\makebox[0pt][c]{%
$^{6}$ Christabel Pankhurst Institute, University of Manchester, UK%
}
\\
\makebox[0pt][c]{%
$^{7}$ NIHR Manchester Biomedical Research Centre, Manchester Academic Health Sciences Centre, Manchester, UK%
}
\\[5pt]
\makebox[0pt][c]{%
\texttt{\small
\{tian.liang, alejandro.frangi, mobarak.hoque, Richard.Bryce\}@manchester.ac.uk}%
}
\\
\makebox[0pt][c]{%
\texttt{\small cjy1125@stanford.edu}%
}
}

\begin{document}

\maketitle
\begingroup
\renewcommand{\thefootnote}{*}
\footnotetext{Corresponding authors}
\endgroup

\begin{abstract}
Early-stage computational drug discovery requires coordinating heterogeneous scientific tools across multi-step workflows. We present a lightweight, tool-augmented framework in which a locally deployable compact language model plans calls to 18 modular tools. A Unified Molecular Schema maintains shared molecular records, while a plug-in interface supports tool replacement and extension. We construct 1,263 manually refined query–plan pairs through workflow-graph path coverage and apply LoRA fine-tuning to three compact model families. Under the query-level split, all fine-tuned models generate fully parseable and schema-compliant plans on 47 held-out cross-group queries. Llama 3.2-3B achieves a tool-selection F1 of 0.998, sequence exact match of 0.979, and argument F1 of 0.960. Under the stricter workflow-grouped split, which excludes identical ordered tool sequences across partitions, sequence exact match reaches 0.452--0.548, highlighting the remaining difficulty for compact models in generating complete workflow paths unseen during training. These results demonstrate the feasibility of compact, locally deployable planning while identifying compositional generalization as an important direction for further improvement.

\end{abstract}

\section{Introduction}
\subsection{Motivation and Contributions}
Early-stage computational drug discovery combines molecular generation, physicochemical and ADMET assessment, novelty evaluation, target-engagement analysis, and synthesis planning \citep{chang2022guide}. These tasks depend on heterogeneous models, databases, and scientific software, while their ordering often changes as intermediate evidence and user-defined constraints become available. Effective workflows therefore require not only accurate individual predictors, but also reliable task orchestration, information exchange, and persistent molecular state across repeated evaluation and prioritization.

General-purpose large language models (LLMs), such as GPT-5.5, Claude Opus 5, and Gemini 3.1 Pro \citep{openai2026gpt55,claude-opus5,gemini31pro}, provide a flexible interface for interpreting scientific requests and coordinating external tools. However, repeatedly invoking remotely hosted models throughout iterative screening can introduce substantial API costs and dependence on model providers and network connectivity. Transmitting tool sequences, filtering criteria, intermediate evaluations, and refinement decisions may also expose project-level research strategies rather than only individual molecular structures. These considerations motivate compact, task-specialized Planners that can perform routine and structurally constrained orchestration on locally controlled infrastructure, while retaining access to specialized external services when required.

Despite the growing use of language-model agents in computational drug discovery, several system-level challenges remain insufficiently addressed. First, early-stage discovery comprises interdependent tasks, yet these capabilities are often integrated as isolated tools without an explicit organization of the underlying task space or their execution dependencies. Second, training data for scientific tool use commonly emphasize individual function calls or manually designed examples, providing limited coverage of the valid multi-step paths that can arise in a realistic discovery workflow. This makes it difficult to train compact planners that can consistently map diverse user requests to appropriate sequences of agents and tools. Furthermore, as molecular candidates move across generation, evaluation, retrieval, and target-analysis stages, different components frequently produce incompatible or loosely structured outputs. Without a unified representation for molecular identity, intermediate evidence, and agent-specific results, information exchange becomes fragile, and the state of a long-running workflow is difficult to maintain. Addressing these gaps requires a systematic organization of early-stage discovery capabilities, workflow-aware supervision for planner specialization, and a shared molecular data structure that supports reliable information flow across the system.

We address these challenges with a lightweight agentic framework in which a fine-tuned compact language model translates user objectives into executable plans across 18 modular tools. A Unified Molecular Schema (UMS) establishes a molecule-centric, block-structured state representation through which heterogeneous tools exchange results across iterative workflows. The framework supports hybrid deployment: routine orchestration and open-source scientific tools can operate locally, while selected capabilities remain accessible through specialized services. Our contributions are: \textbf{(1)} a task-oriented organization and modular tool ecosystem for early-stage drug discovery; \textbf{(2)} a path-coverage-based strategy for constructing supervision and specializing compact Planners; and \textbf{(3)} a shared molecular schema for structured state management across heterogeneous and iterative workflows. Experiments across three model families show that this supervision enables reliable planning with a model as small as 3B parameters in the evaluated tool space.

\subsection{Related Work}

\paragraph{Agentic AI systems for drug discovery.} LLM-based agentic systems have progressed from single-agent tool-use frameworks such as ChemCrow~\citep{bran2024chemcrow} and CoScientist~\citep{boiko2023coscientist} to multi-agent architectures spanning the DMTA cycle, including broad surveys of industrial deployments~\citep{seal2025ai}, end-to-end pipeline proposals~\citep{10.1093/bioadv/vbaf323}, and narrower domain-specific agents for tasks such as ADMET/HTS/DTI prediction~\citep{liu2024drugagent} and pharmaceutical data analysis~\citep{guan2025pharmadata}. However, two aspects remain comparatively underexplored. First, many existing systems rely on general-purpose, remotely hosted LLMs such as GPT-5.5 or Gemini 3.1, while the ability of compact, locally deployable models to perform multi-tool planning has received limited systematic evaluation. Second, workflow knowledge is often encoded through hand-curated tool documentation~\citep{liu2024drugagent}, hard-coded pipeline stages in the system prompt~\citep{guan2025pharmadata}, or ad hoc inter-agent message schemas refined through trial and error~\citep{HE2026104605} rather than as an explicit, machine-checkable dependency structure. Indeed, \citep{HE2026104605} report that migrating AstraZeneca's ChatInvent from a single to a multi-agent architecture improved efficiency but increased tool-call errors, and that resolving these failures required largely ad hoc prompt engineering rather than architectural change, with each LLM upgrade demanding further re-validation. This case illustrates the practical difficulty of maintaining reliable orchestration across architectural and model changes.

\paragraph{Compact models and trajectory-level evaluation.}
Parameter-efficient fine-tuning methods such as LoRA~\citep{hu2022lora} make it practical to specialize compact language models for structured generation and tool use. However, function-calling evaluations commonly emphasize individual tool selection and argument generation, whereas dependency-ordered workflows additionally require the model to compose multiple valid calls in the correct sequence. Evaluation at this level also remains limited. Although BixBench requires agents to conduct long, multi-step analyses, its primary automated evaluation focuses on the final submitted answer rather than the validity of the intermediate tool-calling trajectory~\citep{mitchener2025bixbench}. This distinction is particularly important for scientific orchestration because an invalid intermediate action can propagate errors through subsequent computations even when the final response appears plausible.

Accordingly, we address the orchestration gap by deriving supervision through systematic path coverage over an explicit workflow graph and fine-tuning a compact, locally deployable Planner to generate structured execution plans. We evaluate the Planner directly at the trajectory level, measuring output validity, tool selection, execution ordering, and argument generation, including its ability to compose workflow patterns excluded from training.

\section{System Overview}

\begin{figure}
    \centering
    \includegraphics[width=\textwidth]{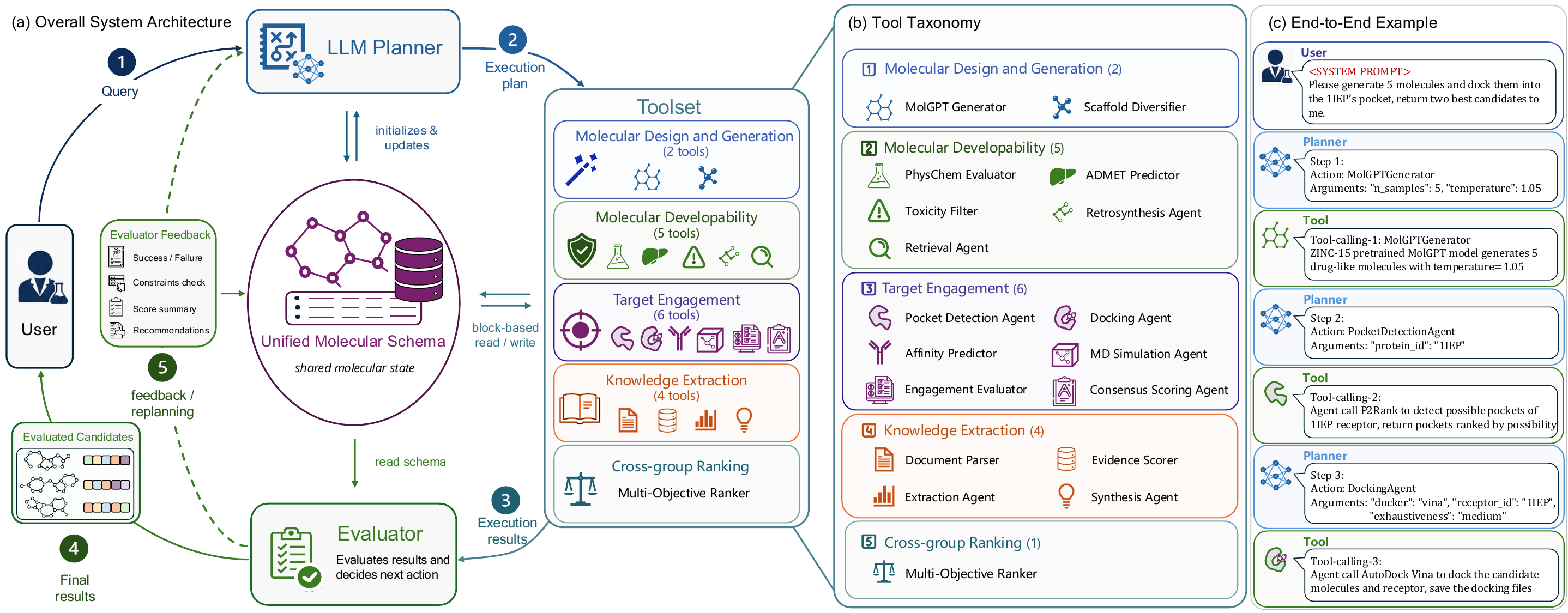}
    \caption{\textbf{(a)} A user query is translated by the fine-tuned \textit{LLM Planner} into an execution plan, which is executed through the toolset. Tools read from and write to authorized blocks of the \textit{Unified Molecular Schema} (UMS), which maintains shared state of each molecule. The \textit{Evaluator} inspects the resulting UMS records and either returns filtered results to the user or provides feedback for replanning to the \textit{Planner} until completion or the retry limit is reached. \textbf{(b)} The current toolset contains 18 tools organized into four functional groups: Molecular Design and Generation, Molecular Developability, Target Engagement, and Knowledge Extraction, together with a cross-group Multi-Objective Ranker. \textbf{(c)} An end-to-end example. Details of this example are shown in the Appendix \ref{sec:dock}.}
    \label{fig1}
\end{figure}

Fig. \ref{fig1} presents the overall architecture of our lightweight, tool-augmented drug discovery system, implemented using LangGraph \citep{wang2024agent}. At its core, the UMS implements a molecule-centric design pattern in which each molecule is maintained as a persistent, incrementally updated record, while tools act as modular producers and consumers of its typed information blocks. These blocks organize information from different computational tasks, including physicochemical properties, ADMET, novelty, target engagement, and synthesis. Rather than passing tool-specific outputs only between adjacent steps, the UMS preserves a shared molecular state across heterogeneous tools and iterative execution cycles (details provided in the Appendix \ref{sec:ums}). Each tool is permitted to read and update only the blocks required by its function, preventing unrelated outputs from being mixed while enabling information exchange through a consistent representation. When external services are required, this block-level access also restricts data transmission to the task-relevant subset of the molecular record, thereby reducing the risk of broad exposure of proprietary project information and intellectual property.

The current system integrates 18 tools in four functional groups: Molecular Design and Generation, Molecular Developability, Target Engagement, and Knowledge Extraction, together with a separate Multi-Objective Ranker for results comparisons. Tools are classified by their primary computational function rather than their chronological position in the drug-development pipeline, since physicochemical, ADMET, and toxicity assessments may also be used during hit discovery. The current toolset comprises both internally implemented or trained tools and integrations of established open-source models and software, with specialized external services used where necessary. Importantly, the architecture is not tied to this particular collection: users may introduce alternative or newly developed tools by adapting their inputs and outputs to the corresponding UMS blocks and execution interface. Detailed descriptions and configurations of the current tools are provided in Appendix \ref{sec:tool_reg}.

A fine-tuned compact language model acts as the \textit{LLM Planner} and can be deployed on locally controlled, GPU-equipped hardware. It converts a user query into an executable sequence of parameterized tool calls, which LangGraph instantiates and runs. Each tool reads the relevant UMS blocks and writes its outputs back to the corresponding blocks. After execution, the Evaluator retrieves the required results from the UMS and determines whether the workflow has produced sufficient information. Successful outputs are filtered by a rule-based reporting layer and returned to the user; otherwise, feedback and execution errors are sent back to the Planner for replanning. This cycle continues until the workflow succeeds or reaches a predefined retry limit, after which a failure report with the corresponding errors is returned.

\section{Lightweight Planner: Data Construction and Evaluation}
\subsection{Planning Task}

\begin{figure}
    \centering
    \includegraphics[width=\textwidth]{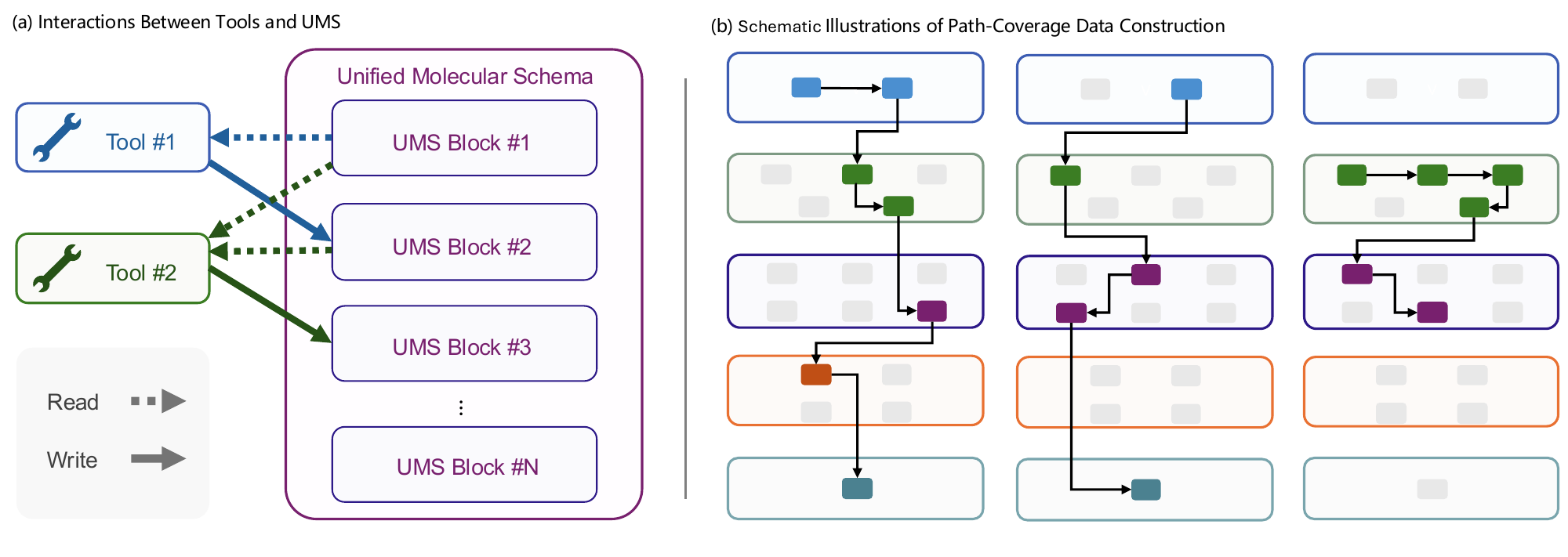}
    \caption{\textbf{(a)} Tools interact with UMS blocks through read (dashed arrows) and write (solid arrows) operations. Each tool can access only its authorized blocks, reducing the risk of unintended data exposure and contamination of molecular records. \textbf{(b)} Three path-coverage workflow examples are illustrated from left to right: a complete workflow spanning molecular generation to ranking and selection of top candidates; a workflow that selects candidates using only a subset of functional layers; and a task-specific workflow that neither begins with molecular generation nor ends with ranking, and may involve more complex sequences within individuals. Colored nodes indicate selected tools, gray nodes indicate unused tools. The functional layers and their color coding are consistent with Fig. \ref{fig1}.}
    \label{fig:data_cons}
\end{figure}

The Planner maps a user query and the available tool definitions to a structured execution plan in JSON format, containing an ordered sequence of parameterized tool calls. Each step specifies a registered tool name and its associated arguments, making both the execution order and the requested operations explicit and machine-readable. The tool definitions describe the available capabilities and argument schemas, allowing the planner to translate the user's objective and constraints into concrete computational steps. During replanning, execution feedback and errors are additionally provided as context. Before execution, each plan is validated for registered tool names, schema-compliant arguments, and dependency-consistent ordering. The selected tools, rather than the Planner, perform scientific computations and update their authorized UMS blocks. The complete output schema and examples are provided in Appendix \ref{sec:output}.

\subsection{Path-Coverage Data Construction}
To construct the fine-tuning dataset, we adopt a workflow-generation strategy inspired by path-coverage testing in software engineering \citep{coverage}. Each tool is represented as a node in a directed graph, and the nodes are arranged according to the functional hierarchy shown in Fig. \ref{fig1}(b). Candidate workflows are then constructed by traversing the graph from higher-level tool groups to lower-level groups. Paths may also connect different tools within the same functional group layer when their operations can be meaningfully combined. Examples of path constructions are shown in Fig. \ref{fig:data_cons}(b).

For each valid path, the ordered tool sequence and its required arguments are defined before natural-language queries are generated. GPT-5.5 \citep{openai2026gpt55} is then prompted to produce multiple semantically diverse user requests whose intended execution corresponds to that path. The model is therefore used to generate linguistic variations rather than to determine the target workflow itself. The resulting queries are manually reviewed and refined to improve clarity, remove ambiguity, and ensure consistency with the associated tool sequence. After invalid or operationally inconsistent paths are removed, this process yields 1,263 query--plan pairs, including 467 cross-group workflows. The dataset generated by GPT-5.5 will undergo manual screening and refinement to improve unnatural phrasing and rewrite queries with substantial overlap.

\subsection{Fine-tuning and Evaluation Setup}

\begin{table*}[t]
\centering
\caption{Planner performance on 47 held-out cross-group queries under the query-level split. Tool-selection, sequence, and argument-generation metrics are averaged over schema-valid outputs. See main text for description of metrics.}
\label{tab:planner_results}
\small
\setlength{\tabcolsep}{5.5pt}
\begin{tabular}{llcccccc}
\toprule
Model & Setting & Parse & Schema & Tool F1 & Seq. EM & Seq. LCS & Arg. F1 \\
\midrule
\multirow{2}{*}{Llama 3.2-3B} & Zero-shot & 0.553 & 0.553 & 0.786 & 0.000 & 0.709 & 0.658 \\
& Fine-tuned & \textbf{1.000} & \textbf{1.000} & \textbf{0.998} & \textbf{0.979} & \textbf{0.996} & \textbf{0.960} \\
\midrule
\multirow{2}{*}{Qwen3-4B} & Zero-shot & 0.298 & 0.298 & 0.778 & 0.000 & 0.716 & 0.736 \\
& Fine-tuned & \textbf{1.000} & \textbf{1.000} & 0.990 & 0.915 & 0.990 & 0.958 \\
\midrule
\multirow{2}{*}{Gemma 3-4B} & Zero-shot & 0.638 & 0.574 & 0.871 & 0.148 & 0.800 & 0.576 \\
& Fine-tuned & \textbf{1.000} & \textbf{1.000} & 0.994 & 0.936 & 0.990 & 0.949 \\
\bottomrule
\end{tabular}
\end{table*}

We fine-tune Llama 3.2-3B, Qwen3-4B, and Gemma 3-4B \citep{grattafiori2024llama, yang2025qwen3, kamath2025gemma} using Low-Rank Adaptation (LoRA) \citep{hu2022lora}. The 1,263 query-plan pairs are divided into training, validation, and test sets using an 80:10:10 ratio. We configure LoRA with a rank of 16, a scaling parameter $\alpha=32$, and a dropout rate of 0.05. The maximum sequence length is set to 4,096 tokens. Training uses an initial learning rate of $2 \times 10^{-4}$ for up to three epochs, and employs early stopping based on validation performance to reduce overfitting. The experiments are conducted on one NVIDIA H200-SXM GPU with 141 GB of memory.

Table \ref{tab:planner_results} reports results on 47 cross-group examples in the query-level test split; complete query-level and workflow-grouped results are provided in Appendix \ref{sub:add}. Fine-tuning improves all models: parse and schema success reach 1.000, while tool selection, ordering, and argument prediction also improve substantially. Llama 3.2-3B performs best overall in this setting. Under the stricter workflow-grouped split, which excludes identical ordered tool sequences across partitions, Seq. EM decreases to 0.452--0.548, showing that generating unseen workflow compositions remains challenging for compact Planners.

\subsection{Case Study}
\label{sec:case}
To illustrate the execution of a cross-group workflow, we asked the system to generate and prioritize candidate molecules against cyclin-dependent kinase 2 (CDK2). The Planner organized four specialized tools: \texttt{MolGPTGenerator} produced 20 candidates, \texttt{PhysChemEvaluator} calculated their physicochemical properties, \texttt{AffinityPredictor} predicted their CDK2 activities, and the rule-based \texttt{MultiObjectiveRanker} ranked the generated molecules according to how well they satisfy the requirements specified in the user query. The resulting records were maintained in the UMS and subsequently prioritized using a target-focused score that assigned 70\% weight to predicted target engagement and 30\% to physicochemical developability in the current version of \texttt{MultiObjectiveRanker}.

The prioritization results are summarized in Table \ref{tab:cdk2_candidates}. Eight of the 20 generated molecules exceeded the predefined potency threshold of predicted pIC$_{50}\geq6.0$. The Rank 1 candidate achieved an overall score of 0.724, with a predicted pIC$_{50}$ of 6.578, a QED of 0.741, and no Lipinski rule-of-five \citep{lipinski1997experimental} violations. Together with the Rank 2 and Rank 3 candidates, it formed a shortlist that balanced predicted CDK2 affinity with physicochemical developability. This case illustrates how the framework coordinates heterogeneous tools, transfers intermediate results through the UMS, and integrates multiple predicted properties into an interpretable ranking. These preliminary results were obtained using the current LightGBM model and may improve with more advanced affinity predictors. Once the evaluation rules are refined, unsatisfactory values will automatically trigger the Evaluator to generate and assess new candidates. We could also note that a pocket-aware molecule generator would likely improve affinity scores. Further details of the case study are provided in Appendix \ref{sec:a_case}.

\begin{table*}[t]
\centering
\caption{Indicative output: top three of twenty candidate ligands generated and evaluated against the protein target, CDK2.}
\label{tab:cdk2_candidates}
\small
\setlength{\tabcolsep}{5pt}
\renewcommand{\arraystretch}{1.15}
\begin{tabularx}{\textwidth}{
    c
    >{\raggedright\arraybackslash}X
    c c c c
}
\toprule
Rank & SMILES & Predicted pIC$_{50}$ & QED & Overall score \\
\midrule
1 &
\texttt{\seqsplit{Cc1nc2ccc(C(=O)N3C[C@H](O)[C@H](NC(=O)[C@H]4COCCO4)C3)cc2n1C}}
& 6.578 & 0.741 & 0.724 \\

2 &
\texttt{\seqsplit{Cn1cc(CC(=O)N2C[C@H](O)[C@H](NC(=O)C[C@H]3CCCOC3)C2)cn1}}
& 6.381 & 0.744 & 0.684 \\

3 &
\texttt{\seqsplit{CCc1cccc(C(=O)NC2(CNC(=O)Cn3cnnn3)CC2)n1}}
& 6.263 & 0.721 & 0.651 \\
\bottomrule
\end{tabularx}
\end{table*}

\section{Conclusion}
We present a lightweight, tool-augmented framework for organizing early-stage computational drug-discovery workflows. The system combines a locally deployable compact Planner with a modular tool ecosystem that is largely built upon open-source scientific software. This design reduces continuous dependence on remotely hosted general-purpose language models while retaining the flexibility to access specialized external services when required. Its plug-in architecture is not tied to the current toolset: individual tools can be replaced or extended as stronger methods become available, provided that their inputs and outputs are aligned with the execution interface and the corresponding Unified Molecular Schema blocks. The UMS further provides persistent, structured molecular records, addressing the challenge of transferring intermediate results across heterogeneous tools and multiple rounds of evaluation and replanning. Experiments across three compact model families show that path-coverage-based supervision can effectively specialize a model as small as 3B parameters for structured workflow orchestration.

The current system nevertheless has several limitations. The compact Planner is specialized for a constrained tool and workflow space and may not generalize reliably to requests, tools, or execution patterns that are insufficiently represented in the training data. Broader deployment will therefore require a larger and more diverse dataset covering additional tool combinations, argument configurations, failure conditions, and replanning scenarios. Close collaboration with drug-discovery researchers and practitioners will be important for constructing realistic user queries and ensuring that future datasets reflect practical scientific workflows rather than only synthetically enumerated paths. In addition, workflow evaluation and result summarization are currently handled by rule-based components. Replacing them with specialized lightweight language models could enable more context-sensitive assessment and reporting, but would require new task-specific datasets and introduce the challenge of deploying multiple models without imposing excessive computational and memory overhead. Finally, although the scientific accuracy of molecular predictions in the current framework is primarily determined by the underlying domain-specific tools and algorithms rather than by the Planner itself, future work should evaluate the system at the level of complete end-to-end workflows. Such evaluation should examine whether tool selection, execution order, state transfer, and iterative replanning collectively preserve the reliability and practical utility of the resulting drug-discovery outcomes.

\section{Acknowledgment}
TL acknowledges Ph.D. funding from the President's Doctoral Scholar Award provided by The University of Manchester. AFF acknowledges support from the Royal Academy of Engineering under the RAEng Chair in Emerging Technologies (INSILEX CiET191919), ERC Advanced Grant-UKRI Frontier Research Guarantee (INSILICO EPY0304941), the UK Centre of Excellence on in-silico Regulatory Science and Innovation (UK CEiRSI) (10139527), the National Institute for Health and Care Research (NIHR) Manchester Biomedical Research Centre (BRC) (NIHR203308), the BHF Manchester Centre of Research Excellence (RE2430017), and the CRUK RadNet Manchester (C1994A28701). The authors would like to acknowledge the assistance given by Research IT and the use of the Computational Shared Facility at the University of Manchester. The authors acknowledge the use of resources provided by the Isambard-AI National AI Research Resource (AIRR) (AIRR-GW-2026-01-01). Isambard-AI is operated by the University of Bristol and is funded by the UK Government’s Department for Science, Innovation and Technology (DSIT) via UK Research and Innovation; and the Science and Technology Facilities Council [ST/AIRR/I-A-I/1023].

% \begin{table}
%   \caption{Sample table caption. Explain what the table shows and add a key take-away message to the caption.}
%   \label{sample-table}
%   \centering
%   \begin{tabular}{lll}
%     \toprule
%     \multicolumn{2}{c}{Part}                   \\
%     \cmidrule(r){1-2}
%     Name     & Description     & Size ($\mu$m) \\
%     \midrule
%     Dendrite & Input terminal  & $\approx$100     \\
%     Axon     & Output terminal & $\approx$10      \\
%     Soma     & Cell body       & up to $10^6$  \\
%     \bottomrule
%   \end{tabular}
% \end{table}

\bibliographystyle{plainnat}
\bibliography{reference}

@article{seal2025ai,
  title={Ai agents in drug discovery},
  author={Seal, Srijit and Huynh, Dinh Long and Chelbi, Moudather and Khosravi, Sara and Kumar, Ankur and Thieme, Mattson and Wilks, Isaac and Davies, Mark and Mustali, Jessica and Sun, Yannick and others},
  journal={arXiv preprint arXiv:2510.27130},
  year={2025}
}

@article{HE2026104605,
title = {Democratising real-world drug discovery through agentic AI},
journal = {Drug Discovery Today},
volume = {31},
number = {2},
pages = {104605},
year = {2026},
issn = {1359-6446},
doi = {https://doi.org/10.1016/j.drudis.2026.104605},
url = {https://www.sciencedirect.com/science/article/pii/S1359644626000103},
author = {Jiazhen He and Helen Lai and Lakshidaa Saigiridharan and Gian Marco Ghiandoni and Kinga Jenei and Umur Gokalp and Ajša Nuković and Ola Engkvist and Jon Paul Janet and Samuel Genheden}
}

@article{10.1093/bioadv/vbaf323,
    author = {Vichentijevikj, Ivana and Mishev, Kostadin and Simjanoska Misheva, Monika},
    title = {Prompt-to-Pill: Multi-Agent Drug Discovery and Clinical Simulation Pipeline},
    journal = {Bioinformatics Advances},
    volume = {6},
    number = {1},
    pages = {vbaf323},
    year = {2026},
    month = {01},
    issn = {2635-0041},
    doi = {10.1093/bioadv/vbaf323},
    url = {https://doi.org/10.1093/bioadv/vbaf323},
    eprint = {https://academic.oup.com/bioinformaticsadvances/article-pdf/6/1/vbaf323/68211589/vbaf323.pdf},
}

@inproceedings{guan2025pharmadata,
  title={PharmaData-Agent: A Specialized Agent for Pharmaceutical Data Analysis},
  author={Guan, Zihan and Wang, Hanyin and Zhou, Zhongliang and Zhou, Qiaohui and Tao, Peining and Ma, Junshui},
  booktitle={2025 IEEE International Conference on Data Mining Workshops (ICDMW)},
  pages={2618--2621},
  year={2025},
  organization={IEEE}
}

@article{liu2024drugagent,
  title={DrugAgent: Automating AI-aided Drug Discovery Programming through LLM Multi-Agent Collaboration},
  author={Liu, Sizhe and Lu, Yizhou and Chen, Siyu and Hu, Xiyang and Zhao, Jieyu and Lu, Yingzhou and Zhao, Yue},
  journal={arXiv preprint arXiv:2411.15692},
  year={2024}
}

@article{bran2024chemcrow,
  title={Augmenting large language models with chemistry tools},
  author={M. Bran, Andres and Cox, Sam and Schilter, Oliver and Baldassari, Carlo and White, Andrew D and Schwaller, Philippe},
  journal={Nature machine intelligence},
  volume={6},
  number={5},
  pages={525--535},
  year={2024},
  publisher={Nature Publishing Group UK London}
}

@article{boiko2023coscientist,
  title={Autonomous chemical research with large language models},
  author={Boiko, Daniil A and MacKnight, Robert and Kline, Ben and Gomes, Gabe},
  journal={Nature},
  volume={624},
  pages={570--578},
  year={2023}
}

@inproceedings{hu2022lora,
  title={LoRA: Low-Rank Adaptation of Large Language Models},
  author={Hu, Edward J and Shen, Yelong and Wallis, Phillip and Allen-Zhu, Zeyuan and Li, Yuanzhi and Wang, Shean and Wang, Lu and Chen, Weizhu},
  booktitle={International Conference on Learning Representations (ICLR)},
  year={2022}
}

@misc{openai2026gpt55,
    author = {{OpenAI}},
    title = {GPT‑5.5 System Card},
    year = {2026},
    month = apr,
    howpublished = {OpenAI},
    note = {Accessed: 2026-07-26}
}

@misc{claude-opus5,
    author = {Anthropic},
    title = {System Card: Claude Opus 5},
    year = {2026},
    month = jul,
    howpublished = {Anthropic},
    note = {Accessed: 2026-07-26}
}

@misc{gemini31pro,
    author = {Google DeepMind},
    title = {Gemini 3.1 Pro Model Card},
    year = {2026},
    month = feb,
    howpublished = {Google DeepMind},
    note = {Accessed: 2026-07-26}
}

@article{grattafiori2024llama,
  title={The llama 3 herd of models},
  author={Grattafiori, Aaron and Dubey, Abhimanyu and Jauhri, Abhinav and Pandey, Abhinav and Kadian, Abhishek and Al-Dahle, Ahmad and Letman, Aiesha and Mathur, Akhil and Schelten, Alan and Vaughan, Alex and others},
  journal={arXiv preprint arXiv:2407.21783},
  year={2024}
}

@article{yang2025qwen3,
  title={Qwen3 technical report},
  author={Yang, An and Li, Anfeng and Yang, Baosong and Zhang, Beichen and Hui, Binyuan and Zheng, Bo and Yu, Bowen and Gao, Chang and Huang, Chengen and Lv, Chenxu and others},
  journal={arXiv preprint arXiv:2505.09388},
  year={2025}
}

@article{kamath2025gemma,
  title={Gemma 3 technical report},
  author={Kamath, Aishwarya and Ferret, Johan and Pathak, Shreya and Vieillard, Nino and Merhej, Ramona and Perrin, Sarah and Matejovicova, Tatiana and Ram{\'e}, Alexandre and Rivi{\`e}re, Morgane and Rouillard, Louis and others},
  journal={arXiv preprint arXiv:2503.19786},
  volume={4},
  year={2025},
  publisher={ArXiv}
}

@article{wang2024agent,
  title={Agent ai with langgraph: A modular framework for enhancing machine translation using large language models},
  author={Wang, Jialin and Duan, Zhihua},
  journal={arXiv preprint arXiv:2412.03801},
  year={2024}
}

@article{coverage,
author = {Zhu, Hong and Hall, Patrick A. V. and May, John H. R.},
title = {Software unit test coverage and adequacy},
year = {1997},
issue_date = {Dec. 1997},
publisher = {Association for Computing Machinery},
address = {New York, NY, USA},
volume = {29},
number = {4},
issn = {0360-0300},
url = {https://doi.org/10.1145/267580.267590},
doi = {10.1145/267580.267590},
journal = {ACM Comput. Surv.},
month = dec,
pages = {366–427},
numpages = {62}
}

@article{lipinski1997experimental,
  title={Experimental and computational approaches to estimate solubility and permeability in drug discovery and development settings},
  author={Lipinski, Christopher A and Lombardo, Franco and Dominy, Beryl W and Feeney, Paul J},
  journal={Advanced drug delivery reviews},
  volume={23},
  number={1-3},
  pages={3--25},
  year={1997},
  publisher={Elsevier}
}

@article{bagal2021molgpt,
  title={MolGPT: molecular generation using a transformer-decoder model},
  author={Bagal, Viraj and Aggarwal, Rishal and Vinod, PK and Priyakumar, U Deva},
  journal={Journal of chemical information and modeling},
  volume={62},
  number={9},
  pages={2064--2076},
  year={2021},
  publisher={ACS Publications}
}

@article{loeffler2024reinvent,
  title={Reinvent 4: Modern AI--driven generative molecule design},
  author={Loeffler, Hannes H and He, Jiazhen and Tibo, Alessandro and Janet, Jon Paul and Voronov, Alexey and Mervin, Lewis H and Engkvist, Ola},
  journal={Journal of Cheminformatics},
  volume={16},
  number={1},
  pages={20},
  year={2024},
  publisher={Springer}
}

@article{moriwaki2018mordred,
  title={Mordred: a molecular descriptor calculator},
  author={Moriwaki, Hirotomo and Tian, Yu-Shi and Kawashita, Norihito and Takagi, Tatsuya},
  journal={Journal of cheminformatics},
  volume={10},
  number={1},
  pages={4},
  year={2018},
  publisher={Springer}
}

@article{fu2024admetlab,
  title={ADMETlab 3.0: an updated comprehensive online ADMET prediction platform enhanced with broader coverage, improved performance, API functionality and decision support},
  author={Fu, Li and Shi, Shaohua and Yi, Jiacai and Wang, Ningning and He, Yuanhang and Wu, Zhenxing and Peng, Jinfu and Deng, Youchao and Wang, Wenxuan and Wu, Chengkun and others},
  journal={Nucleic acids research},
  volume={52},
  number={W1},
  pages={W422--W431},
  year={2024},
  publisher={Oxford University Press}
}

@article{daina2017swissadme,
  title={SwissADME: a free web tool to evaluate pharmacokinetics, drug-likeness and medicinal chemistry friendliness of small molecules},
  author={Daina, Antoine and Michielin, Olivier and Zoete, Vincent},
  journal={Scientific reports},
  volume={7},
  number={1},
  pages={42717},
  year={2017},
  publisher={Nature Publishing Group UK London}
}

@article{sushko2012toxalerts,
  title={ToxAlerts: a web server of structural alerts for toxic chemicals and compounds with potential adverse reactions},
  author={Sushko, Iurii and Salmina, Elena and Potemkin, Vladimir A and Poda, Gennadiy and Tetko, Igor V},
  journal={Journal of chemical information and modeling},
  volume={52},
  number={8},
  pages={2310},
  year={2012}
}

@article{nguyen2021graphdta,
  title={GraphDTA: predicting drug--target binding affinity with graph neural networks},
  author={Nguyen, Thin and Le, Hang and Quinn, Thomas P and Nguyen, Tri and Le, Thuc Duy and Venkatesh, Svetha},
  journal={Bioinformatics},
  volume={37},
  number={8},
  pages={1140--1147},
  year={2021},
  publisher={Oxford University Press}
}

@article{eberhardt2021autodock,
  title={AutoDock Vina 1.2. 0: new docking methods, expanded force field, and python bindings},
  author={Eberhardt, Jerome and Santos-Martins, Diogo and Tillack, Andreas F and Forli, Stefano},
  journal={Journal of chemical information and modeling},
  volume={61},
  number={8},
  pages={3891},
  year={2021}
}

@article{corso2022diffdock,
  title={Diffdock: Diffusion steps, twists, and turns for molecular docking},
  author={Corso, Gabriele and St{\"a}rk, Hannes and Jing, Bowen and Barzilay, Regina and Jaakkola, Tommi},
  journal={arXiv preprint arXiv:2210.01776},
  year={2022}
}

@article{genheden2020aizynthfinder,
  title={AiZynthFinder: a fast, robust and flexible open-source software for retrosynthetic planning},
  author={Genheden, Samuel and Thakkar, Amol and Chadimov{\'a}, Veronika and Reymond, Jean-Louis and Engkvist, Ola and Bjerrum, Esben},
  journal={Journal of cheminformatics},
  volume={12},
  number={1},
  pages={70},
  year={2020},
  publisher={Springer}
}

@article{krivak2018p2rank,
  title={P2Rank: machine learning based tool for rapid and accurate prediction of ligand binding sites from protein structure},
  author={Kriv{\'a}k, Radoslav and Hoksza, David},
  journal={Journal of cheminformatics},
  volume={10},
  number={1},
  pages={39},
  year={2018},
  publisher={Springer}
}

@article{eastman2024openmm,
  title={OpenMM 8: molecular dynamics simulation with machine learning potentials},
  author={Eastman, Peter and Galvelis, Raimondas and Pel{\'a}ez, Ra{\'u}l P and Abreu, Charlles RA and Farr, Stephen E and Gallicchio, Emilio and Gorenko, Anton and Henry, Michael M and Hu, Frank and Huang, Jing and others},
  journal={The Journal of Physical Chemistry B},
  volume={128},
  number={1},
  pages={109--116},
  year={2024},
  publisher={ACS Publications}
}

@article{papadatos2016surechembl,
  title={SureChEMBL: a large-scale, chemically annotated patent document database},
  author={Papadatos, George and Davies, Mark and Dedman, Nathan and Chambers, Jon and Gaulton, Anna and Siddle, James and Koks, Richard and Irvine, Sean A and Pettersson, Joe and Goncharoff, Nicko and others},
  journal={Nucleic acids research},
  volume={44},
  number={D1},
  pages={D1220--D1228},
  year={2016},
  publisher={Oxford University Press}
}

@incollection{bolton2008pubchem,
  title={PubChem: integrated platform of small molecules and biological activities},
  author={Bolton, Evan E and Wang, Yanli and Thiessen, Paul A and Bryant, Stephen H},
  booktitle={Annual reports in computational chemistry},
  volume={4},
  pages={217--241},
  year={2008},
  publisher={Elsevier}
}

@article{europe2015europe,
  title={Europe PMC: a full-text literature database for the life sciences and platform for innovation},
  author={Europe PMC Consortium},
  journal={Nucleic acids research},
  volume={43},
  number={D1},
  pages={D1042--D1048},
  year={2015},
  publisher={Oxford University Press}
}

@article{auer2024docling,
  title={Docling technical report},
  author={Auer, Christoph and Lysak, Maksym and Nassar, Ahmed and Dolfi, Michele and Livathinos, Nikolaos and Vagenas, Panos and Ramis, Cesar Berrospi and Omenetti, Matteo and Lindlbauer, Fabian and Dinkla, Kasper and others},
  journal={arXiv preprint arXiv:2408.09869},
  year={2024}
}

@article{hsu2025llm,
  title={LLM-IE: a python package for biomedical generative information extraction with large language models},
  author={Hsu, Enshuo and Roberts, Kirk},
  journal={JAMIA open},
  volume={8},
  number={2},
  pages={ooaf012},
  year={2025},
  publisher={Oxford University Press}
}

@article{landrum2013rdkit,
  title={Rdkit documentation},
  author={Landrum, Greg and others},
  journal={Release},
  volume={1},
  number={1-79},
  pages={4},
  year={2013}
}

@article{ke2017lightgbm,
  title={Lightgbm: A highly efficient gradient boosting decision tree},
  author={Ke, Guolin and Meng, Qi and Finley, Thomas and Wang, Taifeng and Chen, Wei and Ma, Weidong and Ye, Qiwei and Liu, Tie-Yan},
  journal={Advances in neural information processing systems},
  volume={30},
  year={2017}
}

@article{tu2025askcos,
  title={ASKCOS: open-source, data-driven synthesis planning},
  author={Tu, Zhengkai and Choure, Sourabh J and Fong, Mun Hong and Roh, Jihye and Levin, Itai and Yu, Kevin and Joung, Joonyoung F and Morgan, Nathan and Li, Shih-Cheng and Sun, Xiaoqi and others},
  journal={Accounts of chemical research},
  volume={58},
  number={11},
  pages={1764},
  year={2025}
}

@article{devereux2020extending,
  title={Extending the applicability of the ANI deep learning molecular potential to sulfur and halogens},
  author={Devereux, Christian and Smith, Justin S and Huddleston, Kate K and Barros, Kipton and Zubatyuk, Roman and Isayev, Olexandr and Roitberg, Adrian E},
  journal={Journal of chemical theory and computation},
  volume={16},
  number={7},
  pages={4192--4202},
  year={2020},
  publisher={ACS Publications}
}

@article{chang2022guide,
  title={A guide to in silico drug design},
  author={Chang, Yiqun and Hawkins, Bryson A and Du, Jonathan J and Groundwater, Paul W and Hibbs, David E and Lai, Felcia},
  journal={Pharmaceutics},
  volume={15},
  number={1},
  pages={49},
  year={2022},
  publisher={MDPI}
}

@article{mitchener2025bixbench,
  title={Bixbench: a comprehensive benchmark for llm-based agents in computational biology},
  author={Mitchener, Ludovico and Laurent, Jon M and Andonian, Alex and Tenmann, Benjamin and Narayanan, Siddharth and Wellawatte, Geemi P and White, Andrew and Sani, Lorenzo and Rodriques, Samuel G},
  journal={arXiv preprint arXiv:2503.00096},
  year={2025}
}

@article{batatia2022mace,
  title={MACE: Higher order equivariant message passing neural networks for fast and accurate force fields},
  author={Batatia, Ilyes and Kovacs, David P and Simm, Gregor and Ortner, Christoph and Cs{\'a}nyi, G{\'a}bor},
  journal={Advances in neural information processing systems},
  volume={35},
  pages={11423--11436},
  year={2022}
}

@article{schmidtke2010fpocket,
  title={Fpocket: online tools for protein ensemble pocket detection and tracking},
  author={Schmidtke, Peter and Le Guilloux, Vincent and Maupetit, Julien and Tuffiery, Pierre},
  journal={Nucleic acids research},
  volume={38},
  number={suppl\_2},
  pages={W582--W589},
  year={2010},
  publisher={Oxford University Press}
}

@article{gaulton2012chembl,
  title={ChEMBL: a large-scale bioactivity database for drug discovery},
  author={Gaulton, Anna and Bellis, Louisa J and Bento, A Patricia and Chambers, Jon and Davies, Mark and Hersey, Anne and Light, Yvonne and McGlinchey, Shaun and Michalovich, David and Al-Lazikani, Bissan and others},
  journal={Nucleic acids research},
  volume={40},
  number={D1},
  pages={D1100--D1107},
  year={2012},
  publisher={Oxford University Press}
}

@article{sterling2015zinc,
  title={ZINC 15--ligand discovery for everyone},
  author={Sterling, Teague and Irwin, John J},
  journal={Journal of chemical information and modeling},
  volume={55},
  number={11},
  pages={2324--2337},
  year={2015},
  publisher={ACS Publications}
}

%%%%%%%%%%%%%%%%%%%%%%%%%%%%%%%%%%%%%%%%%%%%%%%%%%%%%%%%%%%%

\appendix
\appendix

\section{System and Tool Specifications}
\subsection{Unified Molecular Schema and State Management}
\label{sec:ums}

\begin{table*}[t]
\centering
\caption{Top-level organization of the Unified Molecular Schema (UMS).
Representative fields are shown; implementation-specific logging and
metadata fields are omitted for brevity.}
\label{tab:ums_structure}
\footnotesize
\setlength{\tabcolsep}{4pt}
\renewcommand{\arraystretch}{1.10}
\begin{tabularx}{\textwidth}{
    >{\raggedright\arraybackslash}p{0.16\textwidth}
    >{\raggedright\arraybackslash}p{0.22\textwidth}
    >{\raggedright\arraybackslash}p{0.27\textwidth}
    >{\raggedright\arraybackslash}X
}
\toprule
Scope & Component & Representative fields & Purpose \\
\midrule

Pipeline
& User objective
& \texttt{user\_query}, \texttt{project\_target},
  \texttt{optimization\_goal}, \texttt{hard\_constraints}
& Stores the original request, optimization objective, and user-defined
workflow constraints. \\

Pipeline
& Target state
& \texttt{target\_id}, \texttt{name}, \texttt{sequence},
  receptor artifacts, target pocket, status
& Maintains target identity, prepared receptor files, detected binding
pockets, and target-preparation status. \\

Pipeline
& Planning and execution
& \texttt{execution\_plan}, \texttt{execution\_trace},
  \texttt{current\_step}, \texttt{replanning\_count},
  \texttt{terminated}
& Records the current plan, execution progress, and information required
for iterative replanning or early termination. \\

Pipeline
& Molecular pool
& \texttt{molecules}, \texttt{molecule\_index},
  \texttt{active\_molecule\_ids}
& Maintains all molecular records, enables identifier-based lookup, and
tracks the active subset passed between tools. \\

Pipeline
& Ranking and nomination
& \texttt{ranking\_strategy}, \texttt{top\_k\_requested},
  \texttt{shortlisted\_ids}, \texttt{nomination\_candidates},
  \texttt{global\_metrics}
& Stores project-level ranking settings, aggregate statistics, and candidate
selection results. \\

Pipeline
& Knowledge state
& \texttt{retrieved\_documents}, \texttt{parsed\_documents},
  \texttt{synthesis\_summary}
& Preserves retrieved and extracted knowledge that is not restricted to
a single molecular record. \\

Pipeline
& Validation and audit
& \texttt{cross\_validation\_matrix},
  \texttt{project\_confidence\_score}, logs, retry requests
& Records cross-tool consistency, project-level confidence, warnings,
failures, and recovery actions. \\

Molecule
& Identity and provenance
& \texttt{molecule\_id}, \texttt{smiles}, \texttt{input\_index},
  \texttt{origin}
& Provides a persistent molecular identity and records whether the molecule
was supplied, generated, or retrieved. \\

Molecule
& Lifecycle state
& \texttt{lifecycle\_stage}
& Tracks progression through \texttt{generated}, \texttt{evaluated},
\texttt{filtered}, \texttt{ranked}, \texttt{shortlisted},
\texttt{nominated}, or \texttt{rejected}. \\

Molecule
& Tool-produced blocks
& \texttt{physchem}, \texttt{affinity}, \texttt{scaffold},
  \texttt{pocket}, \texttt{docking}, \texttt{md}, \texttt{admet},
  \texttt{toxicity}, \texttt{retrosynthesis}, \texttt{novelty}
& Stores heterogeneous scientific results in independent, typed blocks
that can be incrementally populated by authorized tools. \\

Molecule
& Aggregated assessment
& \texttt{consensus\_score}, \texttt{multiobjective\_score},
  \texttt{final\_score}
& Integrates results from multiple property blocks for prioritization and
candidate selection. \\

Molecule
& Flags and history
& \texttt{structural\_alerts}, \texttt{rule\_violations},
  \texttt{evaluation\_flags}, \texttt{evaluation\_history},
  \texttt{artifacts}
& Preserves alerts, validation outcomes, block-update history, and
molecule-specific generated files. \\

\bottomrule
\end{tabularx}
\end{table*}

\begin{table*}[t]
\centering
\caption{Tool-populated blocks within each molecular record in the UMS.
The listed fields are representative rather than exhaustive.}
\label{tab:ums_blocks}
\footnotesize
\setlength{\tabcolsep}{4pt}
\renewcommand{\arraystretch}{1.10}
\begin{tabularx}{\textwidth}{
    >{\raggedright\arraybackslash}p{0.14\textwidth}
    >{\raggedright\arraybackslash}p{0.32\textwidth}
    >{\raggedright\arraybackslash}X
}
\toprule
UMS block & Representative contents & Role in the workflow \\
\midrule

\texttt{physchem}
& Molecular weight, logP, TPSA, hydrogen-bond donors and acceptors,
rotatable bonds, QED, ring statistics, and rule violations
& Describes physicochemical properties and common drug-likeness criteria
used for filtering and multi-objective ranking. \\

\texttt{affinity}
& Target identity, primary endpoint, IC$_{50}$/K$_i$/K$_d$ predictions,
p-values, potency class, threshold outcome, and confidence
& Represents predicted target engagement and provides a normalized potency
value for filtering and ranking. \\

\texttt{scaffold}
& Scaffold SMILES, scaffold type and key, annotation status,
selection status, and removal reason
& Supports scaffold annotation, diversity control, and per-scaffold
candidate quotas. \\

\texttt{pocket}
& Selected pocket, alternative pocket candidates, pocket scores,
probabilities, and docking-box coordinates
& Defines candidate binding sites and provides the search region required
by structure-based docking tools. \\

\texttt{docking}
& Docking method, receptor identifier, execution status, score type,
top score, ranked poses, and pose files
& Stores predicted binding poses and structure-based interaction scores
for downstream assessment or simulation. \\

\texttt{md}
& Simulation duration, mean and standard deviation of RMSD,
binding-energy estimate, and stability decision
& Summarizes molecular-dynamics evidence concerning pose stability and
target engagement. \\

\texttt{admet}
& Absorption, distribution, metabolism, excretion, and toxicity endpoints,
together with probabilities, labels, applicability information, and warnings
& Organizes predicted pharmacokinetic and safety properties for
developability assessment. \\

\texttt{toxicity}
& Filter decision, overall risk, risk level, structural alerts,
toxicophores, reactive groups, and highest-severity finding
& Records rule- or model-based safety assessments and supports hard
safety filtering. \\

\texttt{retrosynthesis}
& Route feasibility, synthetic-accessibility score, route and step counts,
precursors, stock availability, search time, and warnings
& Represents predicted synthetic feasibility and the availability of
candidate synthetic routes. \\

\texttt{novelty}
& Nearest known and patent hits, molecular similarities, structural novelty,
IP novelty, evidence coverage, confidence, and retrieval status
& Summarizes chemical-space novelty and preliminary intellectual-property
risk using evidence from multiple databases. \\

\texttt{artifacts}
& Three-dimensional ligand files, docking input files, generation method,
optimization method, and artifact status
& Associates externally generated files with the corresponding molecular
record while preserving provenance and failure information. \\

\bottomrule
\end{tabularx}
\end{table*}

The UMS separates project-level workflow state from persistent molecule-level records, as summarized in Table \ref{tab:ums_structure}. Each molecular record contains independent tool-populated blocks whose representative contents and workflow roles are listed in Table \ref{tab:ums_blocks}. This block-structured design allows tools to access existing evidence and update only their authorized fields without replacing the complete record. When external services are required, it also restricts data transmission to the task-relevant, authorized subset of the record, thereby reducing the risk of broad exposure of proprietary project information and intellectual property.

\subsection{Tool Registry and Deployment}
\label{sec:tool_reg}

\begin{table*}[t]
\centering
\caption{Registry of molecular design, developability, and target-engagement
tools. Current implementations are those used in the present system;
alternative implementations are illustrative interchangeable back ends and
were not evaluated in this work.}
\label{tab:tool_registry_1}

\fontsize{5.8pt}{6.6pt}\selectfont
\setlength{\tabcolsep}{2.0pt}
\renewcommand{\arraystretch}{0.98}

\begin{adjustbox}{
    max width=\textwidth,
    max totalheight=0.80\textheight,
    center
}
\begin{tabularx}{\textwidth}{
    >{\raggedright\arraybackslash}p{0.105\textwidth}
    >{\raggedright\arraybackslash}p{0.145\textwidth}
    >{\raggedright\arraybackslash}p{0.145\textwidth}
    >{\raggedright\arraybackslash}p{0.135\textwidth}
    >{\raggedright\arraybackslash}X
    >{\raggedright\arraybackslash}p{0.060\textwidth}
}
\toprule
Functional group
& Tool
& Principal inputs
& UMS/state output
& Implementation
& Deployment \\
\midrule

Molecular Design and Generation
& \texttt{MolGPTGenerator}
& Generation size, sampling temperature, and optional molecular constraints
& Molecular identity, SMILES, provenance, and \texttt{generated} lifecycle state
& \textbf{Current:} MolGPT.
  \textit{Alternative:} REINVENT~4 for de novo generation, scaffold decoration,
  linker design, or reinforcement-learning-based optimization
& Local/GPU\\

Molecular Design and Generation
& \texttt{Scaffold-}\newline\texttt{Diversifier}
& Candidate SMILES, ranked identifiers, \texttt{max\_per\_scaffold}, and
  \texttt{top\_k}
& Scaffold annotations and diversity-constrained candidate identifiers
& \textbf{Current:} RDKit Bemis--Murcko scaffold extraction followed by
  per-scaffold selection.
  \textit{Alternative:} fingerprint-based Butina clustering or MaxMin selection
& Local \\

Molecular Developability
& \texttt{PhysChem-}\newline\texttt{Evaluator}
& Molecular SMILES
& \texttt{physchem} block containing MW, logP, TPSA, QED, hydrogen-bond
  counts, and rule violations
& \textbf{Current:} RDKit molecular descriptors.
  \textit{Alternative:} Mordred for an expanded descriptor set
& Local \\

Molecular Developability
& \texttt{ADMET-}\newline\texttt{Predictor}
& Molecular SMILES
& \texttt{admet} block containing predicted absorption, distribution,
  metabolism, excretion, and toxicity endpoints
& \textbf{Current:} ADMETlab~3.0 service.
  \textit{Alternative:} SwissADME for complementary pharmacokinetic and
  drug-likeness assessment
& External \\

Molecular Developability
& \texttt{ToxicityFilter}
& Molecular SMILES and strictness setting
& \texttt{toxicity} block, structural alerts, evaluation flags, and
  filtering decision
& \textbf{Current:} rule-based SMARTS matching.
  \textit{Alternative:} ToxAlerts or RDKit \texttt{FilterCatalog}-based
  alert collections
& Local \\

Molecular Developability
& \texttt{Retrosynthesis-}\newline\texttt{Agent}
& Molecular SMILES, maximum route depth, and route-return setting
& \texttt{retrosynthesis} block containing route feasibility, precursors,
  route steps, and synthetic-accessibility information
& \textbf{Current:} AiZynthFinder.
  \textit{Alternative:} ASKCOS for alternative one-step models and
  multistep synthesis planning
& Local \\

Molecular Developability
& \texttt{RetrievalAgent}
& Molecular structure or search terms, similarity threshold, requested
  databases, and retrieval type
& \texttt{novelty} block, nearest database and patent hits,
  retrieved-document metadata, and source-specific status
& \textbf{Current:} database adapters for ChEMBL and PubChem; DrugBank access
  where licensed; SureChEMBL for patent-associated structures; RDKit
  canonicalization, fingerprints, and Tanimoto similarity; Europe PMC for
  literature metadata and open full text; recursive filesystem lookup for
  user-provided files.
  \textit{Alternative:} locally indexed database mirrors using PostgreSQL
  with the RDKit cartridge
& Hybrid \\

Target Engagement
& \texttt{AffinityPredictor}
& Molecular SMILES and target identifier or sequence
& \texttt{affinity} block containing predicted potency, endpoint,
  confidence, and threshold outcome
& \textbf{Current:} LightGBM-based affinity predictor.
  \textit{Alternative:} GraphDTA or another graph-based drug--target
  affinity model
& Local \\

Target Engagement
& \texttt{PocketDetection-}\newline\texttt{Agent}
& Prepared protein structure and protein identifier
& Target-pocket candidates, ranking scores, probabilities, and docking
  search boxes
& \textbf{Current:} P2Rank~2.5.1.
  \textit{Alternative:} fpocket
& Local \\

Target Engagement
& \texttt{DockingAgent}
& Ligand structure, prepared receptor, binding-site box, docking method,
  and search exhaustiveness
& \texttt{docking} block containing ranked poses, docking scores, status,
  and pose artifacts
& \textbf{Current:} AutoDock Vina for routine docking and DiffDock for
  learning-based pose prediction.
  \textit{Alternative:} GNINA for CNN-assisted docking and rescoring
& Local \\

\bottomrule
\end{tabularx}
\end{adjustbox}
\end{table*}

\begin{table*}[t]
\centering
\caption{Registry of the remaining target-engagement, knowledge-extraction,
and cross-group tools.}
\label{tab:tool_registry_2}

\fontsize{5.8pt}{6.6pt}\selectfont
\setlength{\tabcolsep}{2.0pt}
\renewcommand{\arraystretch}{0.98}

\begin{adjustbox}{
    max width=\textwidth,
    max totalheight=0.80\textheight,
    center
}
\begin{tabularx}{\textwidth}{
    >{\raggedright\arraybackslash}p{0.105\textwidth}
    >{\raggedright\arraybackslash}p{0.145\textwidth}
    >{\raggedright\arraybackslash}p{0.145\textwidth}
    >{\raggedright\arraybackslash}p{0.135\textwidth}
    >{\raggedright\arraybackslash}X
    >{\raggedright\arraybackslash}p{0.060\textwidth}
}
\toprule
Functional group
& Tool
& Principal inputs
& UMS/state output
& Implementation
& Deployment \\
\midrule

Target Engagement
& \texttt{MDSimulation-}\newline\texttt{Agent}
& Protein--ligand complex, simulation duration, and simulation parameters
& \texttt{md} block containing RMSD statistics, stability assessment,
  and optional energetic estimates
& \textbf{Current:} OpenMM with classical molecular-mechanics force fields.
  \textit{Alternative:} OpenMM-ML with ANI-2x or MACE potentials
& Local/GPU \\

Target Engagement
& \texttt{ConsensusScoring-}\newline\texttt{Agent}
& Available affinity, docking, and MD blocks
& Consensus score, component contributions, and combined confidence
& \textbf{Current:} rule-based normalization and weighted aggregation.
  \textit{Alternative:} calibrated stacking or a task-specific meta-model
& Local \\

Target Engagement
& \texttt{Engagement-}\newline\texttt{Evaluation-}\newline\texttt{Agent}
& Target-engagement blocks, confidence values, and execution status
& Conflict flags, cross-validation outcomes, and escalation or retry request
& \textbf{Current:} rule-based consistency and escalation criteria.
  \textit{Alternative:} calibrated conflict classifier or
  natural-language-inference model
& Local \\

Knowledge Extraction
& \texttt{DocumentParser-}\newline\texttt{Agent}
& Local or retrieved PDF, HTML, or XML document
& Structured document object containing sections, tables, figures,
  references, and provenance
& \textbf{Current:} Docling for layout-aware PDF and table parsing,
  supplemented by Beautiful Soup or \texttt{lxml} for HTML/XML.
  \textit{Alternative:} GROBID for scholarly-document structure and metadata
& Local \\

Knowledge Extraction
& \texttt{ExtractionAgent}
& Raw text or structured document object and extraction schema
& Structured experimental claims linking targets, compounds, assays,
  endpoints, quantitative results, and evidence context
& \textbf{Current:} schema-constrained local language-model extraction with
  JSON-schema validation.
  \textit{Alternative:} PubMedBERT/Bioformer-based entity and relation
  extraction or the LLM-IE framework
& Local \\

Knowledge Extraction
& \texttt{EvidenceScoring-}\newline\texttt{Agent}
& Extracted claims and their provenance
& Per-claim confidence, evidence-quality components, limitations, and
  risk-of-bias flags
& \textbf{Current:} rule-based evidence scoring.
  \textit{Alternative:} calibrated evidence classifier or biomedical
  natural-language-inference model
& Local \\

Knowledge Extraction
& \texttt{SynthesisAgent}
& Multiple scored claims and detected conflicts
& Project-level evidence synthesis, consensus statement, unresolved
  conflicts, and confidence
& \textbf{Current:} weighted consensus with rule-based conflict detection.
  \textit{Alternative:} locally deployed domain language model constrained
  to the retrieved and scored evidence
& Local \\

Cross-group
& \texttt{MultiObjective-}\newline\texttt{Ranker}
& Active candidates, available UMS blocks, optimization focus, and
  requested shortlist size
& Per-objective scores, final score, candidate rank, shortlist, and updated
  lifecycle state
& \textbf{Current:} rule-based normalization, constraints, penalties, and
  weighted scoring.
  \textit{Alternative:} Pareto-front or desirability-based ranking,
  calibrated learning-to-rank, or LLM-assisted rationale generation
& Local \\

\bottomrule
\end{tabularx}
\end{adjustbox}
\end{table*}

Table \ref{tab:tool_registry_1} and Table \ref{tab:tool_registry_2} provide the registry of the 18 tools currently defined in the system, including their principal inputs, corresponding UMS or workflow-state outputs, implementation back ends, and deployment requirements. The tools are organized into four functional groups: Molecular Design and Generation, Molecular Developability, Target Engagement, and Knowledge Extraction. The Multi-Objective Ranker is included in the total tool count but is presented separately as a cross-group tool because it integrates results from multiple UMS blocks to prioritize molecular candidates.

The registry distinguishes a tool's scientific function from its underlying implementation. Each tool exposes a standardized execution interface and interacts with typed UMS blocks, allowing its computational back end to be replaced without changing the surrounding workflow. Implementations marked as \textbf{Current} are used in the present system, whereas \textit{Alternative} are included to illustrate compatible replacements or extensions and were not evaluated in this work. Depending on computational requirements and data access, tools may be deployed locally, on GPU-equipped hardware, through external services, or in a hybrid configuration.

The present tool back ends build on MolGPT \citep{bagal2021molgpt}, RDKit \citep{landrum2013rdkit}, ADMETlab 3.0 \citep{fu2024admetlab}, LightGBM \citep{ke2017lightgbm}, AutoDock Vina \citep{eberhardt2021autodock}, DiffDock \citep{corso2022diffdock}, AiZynthFinder \citep{genheden2020aizynthfinder}, P2Rank \citep{krivak2018p2rank}, and OpenMM \citep{eastman2024openmm}. The interchangeable implementations listed in Tables \ref{tab:tool_registry_1} and \ref{tab:tool_registry_2} include REINVENT 4 \citep{loeffler2024reinvent}, Mordred \citep{moriwaki2018mordred}, SwissADME \citep{daina2017swissadme}, ToxAlerts \citep{sushko2012toxalerts}, GraphDTA \citep{nguyen2021graphdta}, ASKCOS \citep{tu2025askcos}, fpocket \citep{schmidtke2010fpocket}, ANI-2x \citep{devereux2020extending}, MACE \citep{batatia2022mace}, Docling \citep{auer2024docling}, and biomedical language models for information extraction \citep{hsu2025llm}. The retrieval layer uses programmatic access to chemical, patent, and literature resources, including ChEMBL \citep{gaulton2012chembl}, PubChem \citep{bolton2008pubchem}, SureChEMBL \citep{papadatos2016surechembl}, and Europe PMC \citep{europe2015europe}.

\section{Planner Dataset Construction and Fine-Tuning}
\subsection{Workflow Graph-Guided Data Construction}
To construct training data for the Planner, we first organize the available tools into a directed workflow graph $G=(V, E)$, where each node $v\in V$ represents a tool and each directed edge $(v_i, v_j)\in E$ indicates that $v_j$ can be validly executed after tool $v_i$. The edges are defined according to the input requirements of each tool, the outputs produced by preceding tools, and the corresponding blocks read from or written to the UMS. For example, a physicochemical property predictor can follow a molecular generation tool because the generated SMILES are stored in the molecular pool, whereas docking requires the necessary ligand and target-structure information to have already been provided or generated. The graph therefore captures executable dependencies between tools rather than imposing a fixed chronological drug-discovery pipeline, allowing workflows to move across functional tool groups according to the user objective.

\begin{figure}[t]
    \centering
    \includegraphics[width=\textwidth]{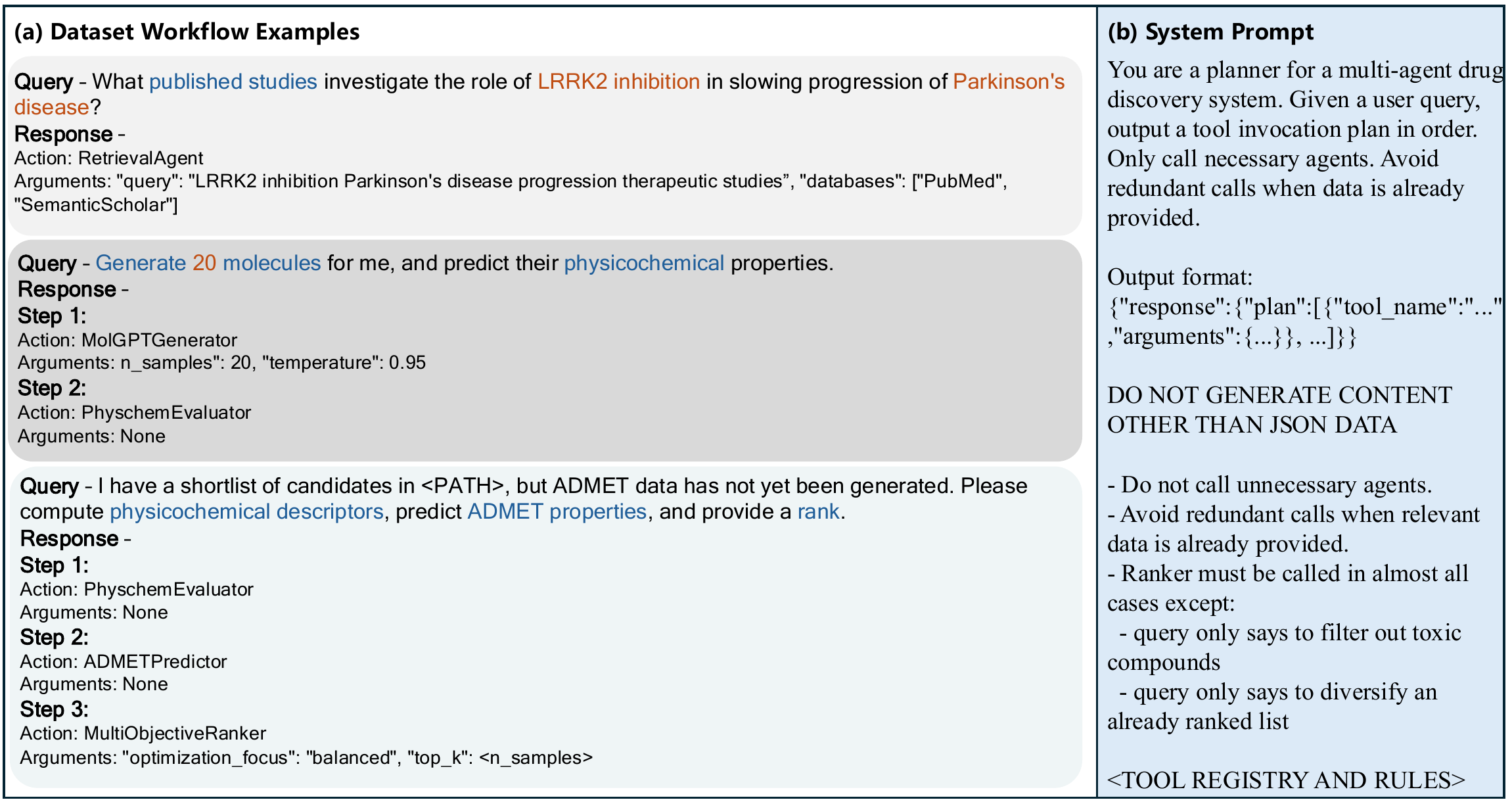}
    \caption{\textbf{(a)} Examples of workflow-guided query--plan pairs in the Planner dataset, with the outer JSONL structure omitted for clarity. Blue text highlights keywords associated with tool calls, while red text indicates keywords corresponding to tool arguments. \textbf{(b)} presents the way to construct the system prompt, while the <TOOL REGISTRY AND RULES> is constructed dynamically according to the tool registry table.}
    \label{fig_dataset}
\end{figure}

Valid tool sequences are then constructed as paths over the workflow graph. We organize the construction process in a hierarchical manner that generally follows a top-down progression across functional group layers, while also allowing intra-group transitions within the same layer. In the top-down setting, paths begin from higher-level functional groups such as molecular generation and progressively proceed to downstream stages including physicochemical evaluation, ADMET prediction, and multi-objective ranking. These paths capture common end-to-end early-stage computational screening workflows and ensure coverage of standard execution pipelines.

In addition to this hierarchical structure, we also construct cross-layer and reverse-direction paths to improve diversity and better reflect real-world usage patterns. Cross-layer paths connect tools across different functional groups. For example, they may link molecular generation with target-affinity prediction, retrosynthesis analysis, or knowledge retrieval. Moreover, certain workflows do not originate from molecular generation; instead, they may start from bottom-layer or intermediate tools such as text-based molecular retrieval, followed by downstream property prediction or optimization. Such reverse or retrieval-initiated paths are particularly important for scenarios where users begin with known compounds or literature-derived molecules rather than de novo generation.

A path is retained only when the required inputs are available from the user query, the initialized system state, or the outputs of preceding tools, thereby excluding incompatible tool combinations and invalid execution orders.

Each selected path is subsequently converted into a structured execution plan containing the ordered tool calls and their required arguments. A large language model is then prompted to generate a natural-language user query whose intent and constraints correspond to the target plan. Query generation is therefore conditioned on a valid workflow, rather than producing an arbitrary query and assigning a tool sequence afterwards. For the current version of the dataset, we used GPT-5.5 \citep{openai2026gpt55} as the model for query generation. This path-to-plan-to-query procedure improves alignment between the natural-language instruction and its supervision signal while allowing variation in molecular targets, input molecules, optimization objectives, constraints, tool arguments, and query phrasing. The resulting query–plan pairs are finally reviewed to remove mismatched intents, invalid tool orderings, malformed arguments, and highly repetitive or unrealistic queries.

As summarized in Table \ref{tab:dataset_content}, the dataset contains 1,263 query-plan pairs, including 232 queries for molecular design and generation, 208 for molecular developability, 185 for target engagement, 171 for knowledge extraction, and 467 cross-group workflows that combine tools from multiple functional groups. The dataset is stored in JSONL format, with each record pairing a natural-language user query with its corresponding structured Planner output. Each execution plan follows a standardized schema consisting of an ordered list of tool calls, where each step specifies an \texttt{action}, its \texttt{arguments}, and a \texttt{step\_id}. For queries that can be answered directly without invoking any tool, the same output format is retained, but the \texttt{plan} field is represented as an empty list. The dataset covers workflows ranging from 0 to 16 steps. Five-step workflows form the largest group, comprising 248 samples (19.64\%), while 573 samples (45.37\%) contain more than five steps. Only 10 samples use an empty plan for queries that require no tool invocation. Fig. \ref{fig_dataset}(a) presents example workflows with the query-plan pair format omitted for clarity.

\subsection{Planner Output Schema and System Prompt}
\label{sec:output}

The Planner is guided by a system prompt that defines its role, the available computational tools, the rules governing tool selection and ordering, and the required output format. Rather than maintaining a separate manually written description of the toolset, the prompt is instantiated from a structured tool registry used by the framework. For the current system, compact descriptions of all 18 available tools are serialized into the prompt at both training and inference time, ensuring that the Planner operates over a consistent and explicitly defined action space.

Each planner-facing registry entry contains only the information required for tool selection and plan construction, including the tool name, its primary function, accepted arguments, required inputs, the UMS blocks it reads from and writes to, and any essential execution prerequisites. Implementation-specific information, such as the underlying model architecture, software dependencies, service endpoints, or internal output fields, is excluded from the Planner prompt. This separation keeps the prompt concise while preserving sufficient information for constructing executable workflows. It also allows the toolset to be extended or modified by updating the structured registry without redesigning the overall prompt format.

The system prompt further instructs the Planner to return only a machine-readable execution plan without additional natural-language explanations. A plan is represented as a variable-length JSON array, in which each element corresponds to one tool call and contains three required fields: \texttt{action}, \texttt{arguments}, and \texttt{step\_id}. The \texttt{action} field must exactly match the name of a registered tool, \texttt{arguments} contains the tool-specific input parameters as key-value pairs, and \texttt{step\_id} specifies the intended execution order. An empty argument object is permitted when a tool requires no explicit parameters because the necessary inputs are already available from the shared system state. The way we construct the system prompt is illustrated in Fig. \ref{fig_dataset}(b).

For example, given the user request, "Generate 20 candidate molecules for CDK2 and evaluate their physicochemical properties and predicted binding affinity", the Planner is expected to produce the execution plan shown in Listing \ref{lst:planner-output}.

\begin{lstlisting}[
    caption={Example Planner output for a CDK2-related query.},
    label={lst:planner-output},
    basicstyle=\ttfamily\small,
    breaklines=true,
    frame=single,
    columns=fullflexible
]
[
  {
    "action": "MolGPTGenerator",
    "arguments": {
      "n_samples": 20,
      "temperature": 1.05
    },
    "step_id": 1
  },
  {
    "action": "PhysChemEvaluator",
    "arguments": {},
    "step_id": 2
  },
  {
    "action": "AffinityPredictor",
    "arguments": {
      "target_id": "CDK2"
    },
    "step_id": 3
  },
  {
      "action": "MultiObjectiveRanker",
      "arguments": {
        "optimization_focus": "target_focused",
        "top_k": 3
      },
      "step_id": 4
    }
]
\end{lstlisting}

Before execution, each generated plan is parsed and checked against both the output schema and the active tool registry. The validation process verifies the JSON structure, confirms that every action refers to a registered tool, checks whether the supplied arguments match the corresponding tool interface, and ensures that the step identifiers specify a coherent execution order. Any plan that fails these checks is rejected before tool execution, preventing invalid Planner outputs from modifying the shared molecular state.

\section{Evaluation Details}
\subsection{Metric Definitions}

The Planner is evaluated in terms of output validity, tool selection, workflow ordering, and argument generation. Let $G_i=(g_{i,1},\ldots,g_{i,m})$ and $P_i=(p_{i,1},\ldots,p_{i,n})$ denote the reference and predicted tool sequences for test query $i$, respectively.

\textbf{Parse Success} is a binary indicator of whether a structured execution plan can be successfully extracted from the model output. \textbf{Schema Success} applies an additional structural check to the extracted plan, verifying that it conforms to the required Planner output schema. In the current schema, a plan is represented as a list of tool-call objects containing the required \texttt{action}, \texttt{arguments}, and \texttt{step\_id} fields.

\texttt{Tool F1} evaluates order-independent tool selection. For each query, the predicted and reference tool sequences are converted into sets, denoted by $\mathcal{P}_i$ and $\mathcal{G}_i$. True positives, false positives, and false negatives are defined as 

\begin{equation}
    TP_i=|\mathcal{P}_i\cap\mathcal{G}_i|,\qquad FP_i=|\mathcal{P}_i\setminus\mathcal{G}_i|, \qquad
    FN_i=|\mathcal{G}_i\setminus\mathcal{P}_i|.
\end{equation}

Tool precision, recall, and F1 are then calculated as

\begin{equation}
    \mathrm{Precision}_i=\frac{TP_i}{TP_i+FP_i},\enspace
    \mathrm{Recall}_i=\frac{TP_i}{TP_i+FN_i},\enspace
    \mathrm{ToolF1}_i=\frac{2\times\mathrm{Precision}_i\times\mathrm{Recall}_i}{\mathrm{Precision}_i+\mathrm{Recall}_i}.
\end{equation}

Workflow ordering is evaluated using \textbf{Sequence Exact Match (Seq. EM)} and \textbf{Sequence Longest Common Subsequence (Seq. LCS)}. Let $\mathbf{1}\{\cdot\}$ denote the indicator function, which returns 1 when the enclosed condition is satisfied and 0 otherwise. Seq. EM is one only when the complete predicted tool sequence is identical to the reference sequence, including its length and ordering:

\begin{equation}
    \mathrm{SeqEM}_i=\mathbf{1}\{P_i=G_i\}.
\end{equation}

Seq. LCS provides partial credit when only part of the reference ordering is preserved. Let $L_i$ be the length of the longest common subsequence between $P_i$ and $G_i$. The score is normalized by the length of the reference sequence:

\begin{equation}
    \mathrm{SeqLCS}_i=\frac{L_i}{|G_i|}.
\end{equation}

\begin{table*}[t]
\centering
\caption{Summary of the hierarchical coverage of the Planner training dataset.}
\label{tab:dataset_content}
\small

\setlength{\tabcolsep}{5pt}
\renewcommand{\arraystretch}{1.15}

\begin{tabularx}{\textwidth}{
    >{\raggedright\arraybackslash}p{0.22\textwidth}
    >{\centering\arraybackslash}p{0.06\textwidth}
    >{\raggedright\arraybackslash}X
}
\toprule
\textbf{Category} & \textbf{Num.} & \textbf{Key Content} \\
\midrule

\makecell[l]{Molecular Design\\and Generation}
&
232
&
De novo molecular generation, scaffold-based diversification, candidate-pool
construction, and control of sampling size and temperature. \\

\makecell[l]{Molecular\\Developability}
&
208
&
Physicochemical property profiling, drug-likeness assessment, ADMET prediction,
toxicity screening, retrosynthetic feasibility analysis, molecular retrieval,
and novelty assessment. \\

\makecell[l]{Target\\Engagement}
&
185
&
Target-specific affinity prediction, binding-pocket detection,
pocket-derived search-box construction, molecular docking,
molecular-dynamics simulation, engagement evaluation, and consensus scoring. \\

\makecell[l]{Knowledge\\Extraction}
&
171
&
Scientific-document parsing, extraction of molecular and target information,
evidence-quality assessment, relation extraction, and synthesis of information
across documents. \\

\makecell[l]{Cross-group\\Workflows}
&
467
&
Workflows spanning multiple functional groups, including generation- or
retrieval-initiated tasks, developability assessment, target-engagement
evaluation, knowledge extraction, and candidate ranking. These workflows
emphasize dependency-aware tool ordering and intermediate-result transfer
through the UMS. \\

\midrule
\textbf{Total}
&
\textbf{1,263}
&
Complete collection of query--plan pairs used for Planner training and
evaluation. \\

\bottomrule
\end{tabularx}
\end{table*}

If the reference sequence is empty, the score is defined as 1 when the predicted sequence is also empty and 0 otherwise. This reference-normalized formulation measures the proportion of the gold-standard workflow that appears in the predicted plan in the correct relative order. Additional predicted steps are not directly penalized by the denominator, although they can affect Tool F1 and Seq. EM.

\textbf{Argument F1 (Arg. F1)} evaluates the arguments assigned to tools that occur in both the predicted and reference plans. Tool calls are first associated by tool name. For each shared tool, the predicted and reference argument dictionaries are compared at the argument-key level. A shared argument key contributes one true positive when its predicted value satisfies the matching rule defined for that tool and argument. A value mismatch contributes both one false positive and one false negative. An additional predicted argument contributes one false positive, whereas an omitted reference argument contributes one false negative. Argument precision, recall, and F1 are calculated from the accumulated counts as

\begin{equation}
    \mathrm{ArgPrecision}_i =
    \frac{TP_i^{\mathrm{arg}}}
    {TP_i^{\mathrm{arg}}+FP_i^{\mathrm{arg}}},
    \qquad
    \mathrm{ArgRecall}_i =
    \frac{TP_i^{\mathrm{arg}}}
    {TP_i^{\mathrm{arg}}+FN_i^{\mathrm{arg}}}.
\end{equation}
\begin{equation}
    \mathrm{ArgF1}_i=\frac{2\times\mathrm{ArgPrecision}_i\times\mathrm{ArgRecall}_i}
    {\mathrm{ArgPrecision}_i+\mathrm{ArgRecall}_i}
\end{equation}

Most argument values are compared exactly, with leading and trailing whitespace removed from string values. Parameter-specific rules are used where small numerical variation or order-independent list comparison is acceptable. For \texttt{MolGPTGenerator}, \texttt{n\_samples} is considered correct when its relative deviation from the reference value does not exceed 30\%, while \texttt{temperature} uses an absolute tolerance of 0.15. The \texttt{similarity\_threshold} argument of \texttt{RetrievalAgent} uses an absolute tolerance of 0.05. Its \texttt{databases} and \texttt{retrieval\_type} arguments are compared as sets, so their ordering and repeated elements are ignored. All remaining arguments use exact matching.

Argument evaluation is conditional on tool selection: only tools appearing in both plans are included in the argument comparison. Consequently, arguments associated with a missing or additional tool call do not contribute directly to Argument F1, because the corresponding tool-selection error is captured separately by Tool F1. The current mapping also assumes that each tool occurs at most once in a plan; if a tool is called repeatedly, only the final occurrence is retained by the tool-to-step mapping.

\subsection{Additional Results and Error Analysis}
\label{sub:add}

\subsubsection{Overall Performance on the Full Dataset}

\begin{table*}[t]
\centering
\caption{Planner performance on the full held-out test split ($N=127$). Llama, Qwen, and Gemma denote Llama 3.2-3B, Qwen3-4B, and Gemma 3-4B, respectively. ZS and F denote zero-shot and fine-tuned settings. P and R denote precision and recall. Tool-selection, sequence, and argument-generation metrics are averaged over schema-valid outputs. Bold indicates the highest displayed value in each column; ties after rounding are both highlighted.}
\label{tab:additional-results}
\scriptsize
\setlength{\tabcolsep}{3.5pt}
\renewcommand{\arraystretch}{1.08}
\begin{tabular}{llcccccccccc}
\toprule
Model & Set. & \multicolumn{2}{c}{Validity} & \multicolumn{3}{c}{Tool selection} & \multicolumn{2}{c}{Sequence} & \multicolumn{3}{c}{Argument generation} \\
\cmidrule(lr){3-4}\cmidrule(lr){5-7}\cmidrule(lr){8-9}\cmidrule(lr){10-12}
& & Parse & Schema & P & R & F1 & EM & LCS & P & R & F1 \\
\midrule
\multirow{2}{*}{Llama} & ZS & 0.606 & 0.606 & 0.669 & 0.734 & 0.666 & 0.065 & 0.670 & 0.512 & 0.576 & 0.539 \\
& F & \textbf{1.000} & \textbf{1.000} & \textbf{0.989} & \textbf{0.989} & \textbf{0.988} & \textbf{0.961} & \textbf{0.997} & \textbf{0.936} & \textbf{0.942} & \textbf{0.939} \\
\midrule
\multirow{2}{*}{Qwen} & ZS & 0.591 & 0.591 & 0.666 & 0.718 & 0.662 & 0.080 & 0.651 & 0.515 & 0.568 & 0.537 \\
& F & \textbf{1.000} & \textbf{1.000} & \textbf{0.989} & 0.984 & 0.986 & 0.937 & 0.991 & 0.934 & 0.941 & 0.937 \\
\midrule
\multirow{2}{*}{Gemma} & ZS & 0.803 & 0.740 & 0.680 & 0.779 & 0.703 & 0.085 & 0.732 & 0.618 & 0.571 & 0.584 \\
& F & \textbf{1.000} & \textbf{1.000} & 0.984 & 0.986 & 0.982 & 0.929 & 0.989 & 0.926 & 0.930 & 0.928 \\
\bottomrule
\end{tabular}
\end{table*}

Table \ref{tab:additional-results} reports the results on the complete held-out test set of 127 queries. Fine-tuning substantially improved both output validity and planning accuracy across all three model families. After fine-tuning, every model achieved parse and schema success rates of 1.000, indicating that all generated outputs could be extracted and conformed to the required Planner output structure. Llama 3.2-3B achieved the strongest overall performance, with a Tool F1 of 0.988, a sequence exact-match rate of 0.961, a Sequence LCS score of 0.997, and an Argument F1 of 0.939. Qwen3-4B and Gemma-3-4B performed similarly after fine-tuning, although their sequence exact-match rates were slightly lower at 0.937 and 0.929, respectively. The consistently high Sequence LCS scores, ranging from 0.989 to 0.997, show that even when a predicted plan was not an exact match, its tool ordering generally remained close to the reference workflow.

The zero-shot models showed considerably lower output validity and planning accuracy. Llama 3.2-3B and Qwen3-4B produced schema-valid plans for only 60.6\% and 59.1\% of the test queries, respectively. Gemma-3-4B showed higher zero-shot parse and schema success rates of 80.3\% and 74.0\%, but its sequence exact-match rate remained only 0.085. The difference between its parse and schema success rates indicates that some outputs contained extractable structured plans but lacked required schema elements, most commonly the \texttt{arguments} field in one or more steps. Thus, the ability to produce parseable JSON did not necessarily imply compliance with the executable Planner output schema. Across all three models, zero-shot sequence exact-match rates remained below 0.09, demonstrating that general instruction-following ability alone was insufficient to reliably reproduce the expected tool composition, ordering, and configuration.

Tool-selection, sequence, and argument-generation metrics are calculated only over schema-valid outputs. This conditional denominator is used because these metrics require an explicitly structured plan in which the \texttt{action}, \texttt{arguments}, and \texttt{step\_id} fields can be consistently identified. For an output that cannot be parsed or does not satisfy the schema, the predicted tool sequence and argument dictionaries are undefined or structurally incomplete, making direct tool-, order-, and argument-level comparison unreliable. Restricting these metrics to schema-valid outputs therefore measures semantic planning quality after the model has produced an evaluable plan, whereas Parse Success and Schema Success separately quantify how often the model reaches this evaluable state. Reporting both groups of metrics is necessary: the conditional planning scores should not be interpreted as success rates over all 127 queries, particularly for the zero-shot models. Assigning a score of zero to every structurally invalid output would instead combine formatting failure and planning error within the same metric and would duplicate failures already captured by the validity measures.

A detailed analysis of Llama-3.2-3B further shows that fine-tuning changed the nature of the remaining errors. The fine-tuned model omitted no required arguments, and most residual errors involved incorrect values rather than structural failures. The most common cases concerned the number of candidates requested from the multi-objective ranker (12 cases), the retrieval query (7), and the target identifier supplied for affinity prediction (5). Unsupported additional arguments were rare, occurring only three times in the ranker and twice in the docking configuration.

The zero-shot model, by contrast, exhibited a broader range of configuration errors. These frequently affected the number of ranked candidates (18 cases), the number of molecules to generate (17), the generation temperature (14), the molecular-dynamics simulation duration (12), and the toxicity-filtering mode (12). It also supplied an unsupported candidate-source indicator to the ranker in 28 cases and occasionally omitted required retrieval and docking parameters.

These findings suggest that fine-tuning improved not only structural compliance and workflow construction, but also the translation of query-specific entities, numerical constraints, and configuration requirements into executable arguments. The remaining errors of the fine-tuned Planner were therefore concentrated mainly in exact numerical values and target-specific identifiers, rather than in the overall workflow structure.

\subsubsection{Generalization under the Workflow-Grouped Split}
To better understand whether the strong fine-tuning performance is driven by true generalization or by the characteristics of the dataset split, we further conducted a workflow-grouped split analysis. In the original dataset, samples were partitioned randomly at the query level, which means that different natural language variations describing the same underlying workflow could be distributed across training and test sets. This setup may inadvertently make the evaluation easier, as the model could implicitly benefit from seeing semantically similar or structurally equivalent workflows during training, even if the surface-form queries differ. To control for this potential leakage of workflow-level information, we introduce a stricter evaluation setting in which all samples sharing the same ordered tool sequence are assigned to the same partition, ensuring that no identical ordered tool sequence appears in more than one data partition.

\begin{table*}[t]
\centering
\caption{Planner performance on the workflow-grouped test split
($N=126$). Samples sharing the same ordered tool sequence are assigned
to the same data partition. Llama, Qwen, and Gemma denote Llama-3.2-3B,
Qwen3-4B, and Gemma-3-4B, respectively. ZS and F denote
zero-shot and fine-tuned settings. P and R denote precision and
recall. Parse and Schema are calculated over all 126 test samples;
all remaining metrics are averaged only over schema-valid outputs.
The corresponding valid sample counts are 70, 82, and 93 for the
three zero-shot models and 126 for each fine-tuned model. Bold
indicates the best fine-tuned result in each column.}
\label{tab:grouped-results}
\scriptsize
\setlength{\tabcolsep}{3.5pt}
\renewcommand{\arraystretch}{1.08}
\begin{tabular}{llcccccccccc}
\toprule
Model & Set. &
\multicolumn{2}{c}{Validity} &
\multicolumn{3}{c}{Tool selection} &
\multicolumn{2}{c}{Sequence} &
\multicolumn{3}{c}{Argument generation} \\
\cmidrule(lr){3-4}
\cmidrule(lr){5-7}
\cmidrule(lr){8-9}
\cmidrule(lr){10-12}
& & Parse & Schema & P & R & F1 & EM & LCS & P & R & F1 \\
\midrule
\multirow{2}{*}{Llama}
& ZS & 0.556 & 0.556 & 0.790 & 0.793 & 0.749
& 0.129 & 0.763 & 0.626 & 0.672 & 0.645 \\
& F & 1.000 & 1.000 & 0.955 & 0.879 & 0.909
& \textbf{0.548} & 0.865 & 0.901 & 0.907 & 0.904 \\
\midrule
\multirow{2}{*}{Qwen}
& ZS & 0.651 & 0.651 & 0.681 & 0.685 & 0.661
& 0.110 & 0.612 & 0.489 & 0.555 & 0.518 \\
& F & 1.000 & 1.000 & 0.966 & \textbf{0.896} & \textbf{0.926}
& 0.508 & \textbf{0.889} & \textbf{0.910}
& \textbf{0.913} & \textbf{0.911} \\
\midrule
\multirow{2}{*}{Gemma}
& ZS & 0.810 & 0.738 & 0.758 & 0.788 & 0.761
& 0.032 & 0.707 & 0.596 & 0.480 & 0.528 \\
& F & 1.000 & 1.000 & \textbf{0.969} & 0.865 & 0.910
& 0.452 & 0.857 & 0.881 & 0.880 & 0.881 \\
\bottomrule
\end{tabular}
\end{table*}

As shown in Table \ref{tab:grouped-results}, all fine-tuned models retained perfect parse and schema success rates. Qwen3-4B achieved the highest Tool F1 (0.926), Sequence LCS (0.889), and Argument F1 (0.911), while Llama 3.2-3B obtained the highest Sequence EM (0.548). Although exact workflow reproduction decreased under this more challenging split, the substantially higher LCS scores indicate that most required tools were still generated in the correct relative order.

Zero-shot performance remained substantially lower than fine-tuned performance. Llama 3.2-3B achieved the lowest parse and schema success rates (0.556), with 56 outputs failing plan extraction, but performed relatively well on 70 valid outputs, achieving a Sequence EM of 0.129, a Sequence LCS of 0.763, and an argument F1 of 0.645. Gemma 3-4B produced the highest output validity and Tool F1 (0.761), although some parsed plans omitted the required \texttt{arguments} field. Overall, fine-tuning consistently improved structural validity, tool selection, workflow ordering, and argument generation across all three models.

\paragraph{Overall discussion.}
The substantial gains under the query-level split demonstrate the effectiveness of fine-tuning for mapping new query formulations to structured execution plans. However, because different queries derived from the same ordered workflow can occur across data partitions, the strongest results should not be interpreted as generalization to entirely unseen workflow compositions. Under the stricter workflow-grouped split, Seq. EM decreases to 0.452--0.548 across the three fine-tuned models, showing that compositional generalization remains challenging. However, this experiment also provides evidence of the Planner's robustness: for workflows represented in the training set, it continues to generate accurate plans despite variations in query phrasing. The perfect parse and schema success rates may also partly reflect the explicit output-format instructions, while the detailed tool descriptions provide strong guidance for tool selection and argument generation. Moreover, the dataset remains relatively small, and most queries state the desired operations directly, with limited coverage of ambiguous or context-dependent requests. Future work will therefore expand the dataset with practitioner-informed, implicitly expressed, incomplete, and domain-specific instructions, together with broader coverage of held-out workflow compositions.

\section{Case Study Details}
\subsection{CDK2 Affinity-Based Candidate Prioritization}
\label{sec:a_case}
\subsubsection{Case Study Setup}
We constructed a CDK2-targeted molecular design case study in Section \ref{sec:case} to demonstrate how the framework translates a natural-language discovery objective into an executable workflow and maintains molecule-level results across successive computational tools. The input task requested the generation of 20 candidate molecules, followed by evaluation of their physicochemical properties and prediction of their affinity toward CDK2, and prioritization of the top three candidates. This setting represents an early-stage virtual screening scenario in which newly generated candidates must be assessed jointly for molecular developability and target engagement.

For this case study, the Planner received the following query: \textit{"Generate 20 candidate molecules, compare their physicochemical properties and predicted affinities toward CDK2, and return the three highest-ranked candidates."} The Planner, a fine-tuned Llama 3.2-3B model, produced a four-step workflow demonstrated in Listing \ref{lst:planner-output}. The molecule generator was performed using a version of MolGPT \citep{bagal2021molgpt} pretrained on the ZINC15 dataset \citep{sterling2015zinc}, configured with \texttt{n\_samples = 20} and \texttt{temperature = 1.05}. Affinity prediction was performed using \texttt{target\_id = "CDK2"}. For this step, compounds with experimentally measured $\mathrm{IC}_{50}$ values for CDK2 were retrieved from ChEMBL (ChEMBL Target ID: CHEMBL301). Molecular features derived from their SMILES were paired with the corresponding converted $\mathrm{pIC}_{50}$ values to train a target-specific LightGBM \citep{ke2017lightgbm} regression model. The trained LightGBM model was then used by \texttt{AffinityPredictor} to predict the $\mathrm{pIC}_{50}$ values of the generated candidates against CDK2. Each generated molecule was assigned a unique identifier in the UMS, enabling the physicochemical and affinity results produced by subsequent tools to be associated with the same molecular record.

As the fourth step of the Planner-generated workflow, \texttt{MultiObjectiveRanker} prioritized the evaluated candidates using a rule-based configuration that assigned 70\% of the composite score to predicted CDK2 affinity and 30\% to physicochemical developability, with additional penalties applied for molecular liabilities. Predicted affinity was evaluated on an absolute $\mathrm{pIC}_{50}$ scale rather than being normalized relative to the generated batch, and $\mathrm{pIC}_{50}\geq6.0$ was used as the potency threshold.

\subsubsection{Workflow Execution and Candidate Prioritization}
During execution, \texttt{MolGPTGenerator} produced 20 candidate molecules, which were registered as individual records in the UMS. For each candidate, \texttt{PhysChemEvaluator} calculated the physicochemical descriptors and wrote them directly to the \texttt{physchem} block of the corresponding molecular record. \texttt{AffinityPredictor} then predicted activity toward CDK2 and stored the result in the \texttt{affinity} block of the same record. Associating both outputs with the same \texttt{molecule\_id} allowed the ranking procedure to retrieve and jointly evaluate the physicochemical and affinity information directly from the UMS, without requiring separate tool-specific data structures or intermediate data conversion.

Once both blocks had been populated, \texttt{MultiObjectiveRanker} retrieved the affinity and physicochemical results directly from each molecular record and calculated the composite score using the configuration described above. The affinity component was derived from the absolute predicted $\mathrm{pIC}_{50}$, while the developability component summarized the physicochemical profile and incorporated penalties for identified molecular liabilities. Because affinity was not normalized against the other molecules in the generated batch, the score assigned to a candidate did not depend on the affinity range of the remaining candidates. The threshold $\mathrm{pIC}_{50}\geq6.0$ was used to identify candidates with predicted submicromolar activity rather than to determine their ordering solely by potency.

Eight of the 20 generated molecules satisfied this potency threshold. The highest-ranked candidate achieved a composite score of 0.724, a predicted $\mathrm{pIC}_50$ of 6.578, and a QED score of 0.741, with no Lipinski Rule-of-Five violations. The corresponding $\mathrm{IC}_{50}$ was calculated as $10^{9-\mathrm{pIC}_{50}}$ nM, giving approximately 264.06 nM. Some candidates with marginally higher predicted affinity received lower overall scores because of less favorable physicochemical profiles or additional liability penalties. The final ordering therefore reflects a balance between predicted target engagement and developability rather than a potency-only ranking. As all affinity and property values were computationally estimated, the ranking represents candidate prioritization within the framework and not experimentally validated CDK2 inhibition.

\subsection{Structure-Based Docking with PDB 1IEP}
\label{sec:dock}
\subsubsection{Case Study Setup and Receptor Preparation}
This case study evaluates whether the framework can coordinate molecular generation, binding-pocket detection, structure-based docking, and candidate prioritization within a single workflow. The Planner received the following query: \textit{"Generate five candidate molecules, identify a binding pocket in the 1IEP receptor structure, dock the generated molecules into the detected pocket, and return the two highest-ranked candidates."} The Planner generated the workflow shown in Listing \ref{lst:planner-output-1iep}. \texttt{PocketDetectionAgent} uses P2Rank version 2.5.1, and was configured to return the three highest-ranked pockets using four CPU threads. When \texttt{docker="vina"}, \texttt{DockingAgent} performs molecular docking using AutoDock Vina, implemented via its Python package (version 1.2.7).

\begin{lstlisting}[
    caption={Planner output for docking-based candidate prioritization.},
    label={lst:planner-output-1iep},
    basicstyle=\ttfamily\small,
    breaklines=true,
    frame=single,
    columns=fullflexible
]
[
  {
    "action": "MolGPTGenerator",
    "arguments": {
      "n_samples": 5,
      "temperature": 1.05
    },
    "step_id": 1
  },
  {
    "action": "PocketDetectionAgent",
    "arguments": {
      "protein_id": "1IEP"
    },
    "step_id": 2
  },
  {
    "action": "DockingAgent",
    "arguments": {
      "docker": "vina",
      "receptor_id": "1IEP",
      "exhaustiveness": "medium"
    },
    "step_id": 3
  },
  {
      "action": "MultiObjectiveRanker",
      "arguments": {
        "optimization_focus": "balanced",
        "top_k": 3
      },
      "step_id": 4
    }
]
\end{lstlisting}

\begin{figure}[!h]
    \centering
    \includegraphics[width=\textwidth]{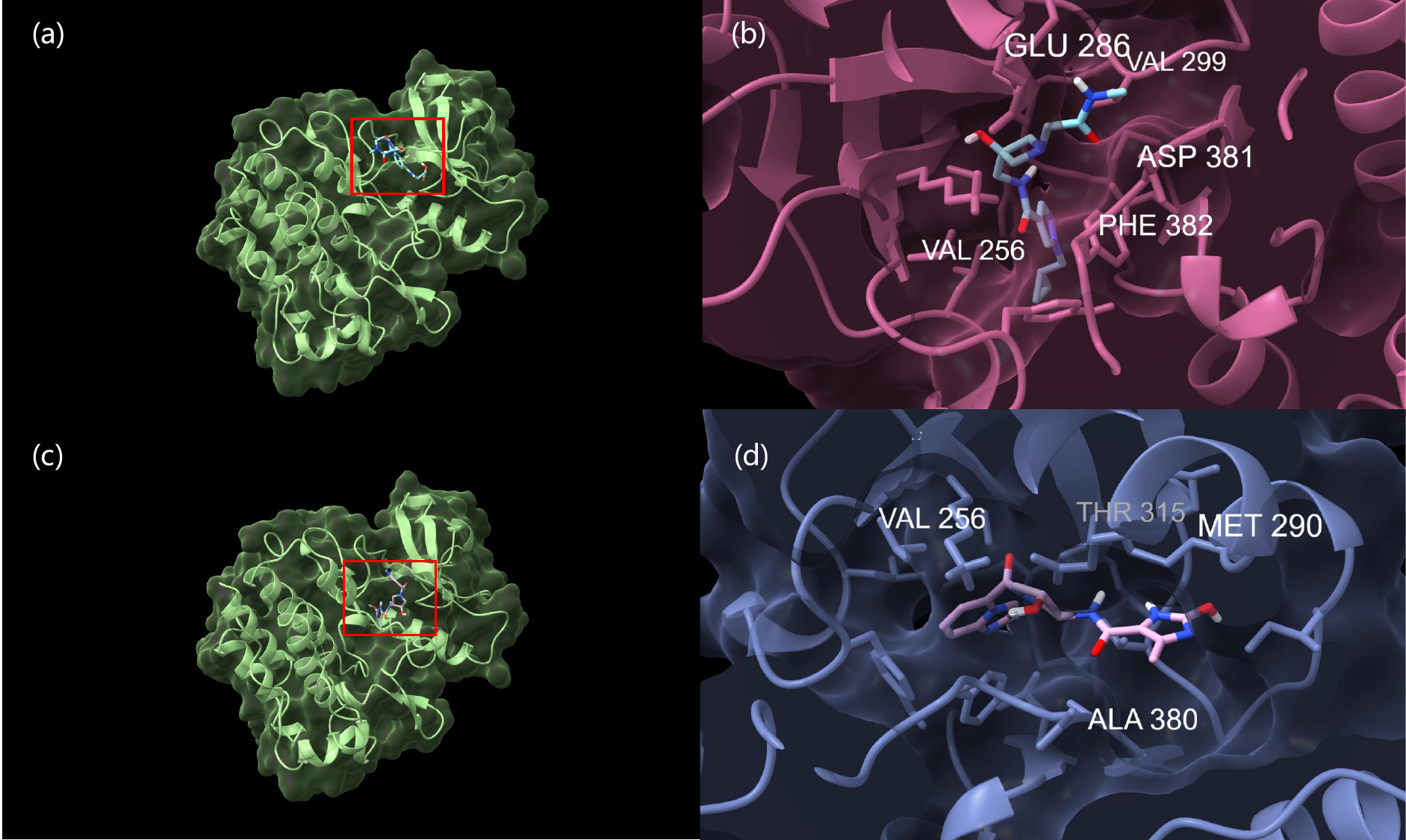}
    \caption{Predicted binding poses of the two top-ranked candidates in the selected pocket of the c-Abl kinase protein. Panels \textbf{(a)} and \textbf{(b)} show the overall and close-up views of \texttt{mol1}, respectively, with a Vina score of -9.8 kcal/mol; panels \textbf{(c)} and \textbf{(d)} present the corresponding views of \texttt{mol2}, with a Vina score of -8.5 kcal/mol. Red boxes indicate the enlarged binding regions, surrounding residues are labeled.}
    \label{fig2}
\end{figure}

The receptor was derived from PDB entry 1IEP (pdb\_00001iep), a 2.10 \AA{} X-ray crystal structure of the c-Abl kinase domain in complex with small molecule ligand STI-571. A preprocessed receptor with hydrogen atoms assigned file, \texttt{1iep\_receptorH.pdb}, was provided through the initialized target state rather than downloaded during the workflow execution. This PDB representation was used as the input to \texttt{PocketDetectionAgent}, whereas \texttt{DockingAgent} generated or retrieved the corresponding Vina-compatible receptor representation before docking.

Binding-pocket detection was performed using P2Rank version 2.5.1. The predicted pockets were normalized into the UMS pocket block, including their rank, score, probability, center coordinates, and spatial extent. For each pocket, the system constructed a \texttt{VinaBox} containing the box center and dimensions required by AutoDock Vina. The highest-ranked predicted pocket was selected for docking, after which the generated ligands were converted into three-dimensional Vina-compatible representations and docked within the resulting search box.

\subsubsection{Pocket Detection and Docking Workflow}
The workflow shown in Listing \ref{lst:planner-output-1iep} first established the ligand-side and receptor-side inputs required for docking. \texttt{MolGPTGenerator} generated five candidate ligands and stored their SMILES representations in the molecular pool, while \texttt{PocketDetectionAgent} analyzed the initialized 1IEP receptor structure using P2Rank and retained the three highest-ranked predicted pockets. Although these tools were executed sequentially, they provided independent and complementary inputs rather than forming a direct dependency. The top-ranked pocket was selected for docking, and its center coordinates and spatial extent were converted into the \texttt{VinaBox} representation required by \texttt{DockingAgent}.

For each active candidate molecule, \texttt{DockingAgent} generated a three-dimensional ligand structure, converted it into a Vina-compatible representation, and docked it into the selected pocket using the prepared receptor and pocket-derived search box. The predicted binding affinities and docked poses were written to the corresponding docking block and artifact records in the UMS. Finally, \texttt{MultiObjectiveRanker} read the available docking results and prioritized the successfully docked candidates according to the configured docking-based ranking criterion, with more negative Vina binding affinities indicating more favorable predicted binding. Candidates for which ligand preparation or docking failed were excluded from the final ranking.

\subsubsection{Docking Results and Candidate Prioritization}

P2Rank identified the highest-ranked pocket with a pocket score of 16.26 and a probability of 0.774. The pocket was successfully converted into a docking-ready \texttt{VinaBox} and used for docking all generated candidate molecules.

Based on the resulting docking scores, \texttt{MultiObjectiveRanker} returned the two candidates with the most favorable predicted binding energies. The top-ranked candidate, \texttt{mol1}, achieved a Vina score of -9.8 kcal/mol, while the second-ranked candidate, \texttt{mol2}, achieved a score of -8.5 kcal/mol. Their molecular structures were represented by the following SMILES:

\begin{itemize}
\item \texttt{mol1}: \nolinkurl{C[C@H]1C(=O)N(C)CCN1C(=O)C(=O)Nc1ccc(C(=O)NCCC(=O)O)cc1}
\item \texttt{mol2}: \nolinkurl{Cc1nc(O)[nH]c1C(=O)N[C@@H]1CN(C(=O)c2cccc3nccn23)C[C@H]1O}
\end{itemize}

Fig. \ref{fig2} presents the predicted binding poses of the two candidates. For \texttt{mol1}, the docked pose is positioned near residues including Val256, Glu286, Val299, Asp381, and Phe382, with the close-up view indicating possible polar contacts within the pocket. The predicted pose of \texttt{mol2} is surrounded by residues including Val256, Met290, Thr 315, and Ala380. These visualizations illustrate how the workflow connects pocket detection, docking, and candidate prioritization through the shared molecular state. However, the reported scores and poses are computational predictions and should not be interpreted as experimental evidence of binding to c-Abl.

\end{document}